\documentclass[journal]{IEEEtran}
\usepackage{amsmath,amsfonts}

\usepackage{algorithm}

\usepackage{algpseudocode}
\usepackage{array}
\usepackage[caption=false,font=normalsize,labelfont=sf,textfont=sf]{subfig}
\usepackage{textcomp}
\usepackage{stfloats}
\usepackage{url}
\usepackage{verbatim}
\usepackage{graphicx}
\usepackage{cite}
\usepackage{booktabs}
\usepackage[table]{xcolor}
  
\newcommand{\shline}{\midrule}
\newcolumntype{x}[1]{>{\centering\arraybackslash}p{#1pt}}
\newcolumntype{y}[1]{>{\centering\arraybackslash}p{#1pt}}
\usepackage{multirow}
\usepackage{bbding}
\usepackage{pifont}

\usepackage[colorlinks=true,citecolor=blue,linkcolor=blue,urlcolor=black]{hyperref}

\begin{document}

\title{Multi-Modal Object Re-Identification with Dual Semantic Guidance and Global-Local Mutual Modulation}

\author{Weixiang Zhou, Xingguo Xu, Yuhao Wang, Cong Wang, Yang Yang, Zhixun Su,~\IEEEmembership{Member,~IEEE,} and Jinshan~Pan,~\IEEEmembership{Senior~Member,~IEEE}

\thanks{ This work was supported in part by the National Natural Science Foundation of China under Grant 62476041. \textit{(Weixiang Zhou and Xingguo Xu contributed equally to this work.) (Corresponding author: Zhixun Su.)}}

\thanks{
Weixiang Zhou, Xingguo Xu, and Zhixun Su are with the School of Mathematical Sciences and the Key Laboratory for Computational Mathematics and Data Intelligence of Liaoning Province, Dalian University
of Technology, Dalian, 116024, China (e-mail: s20201162006@mail.dlut.edu.cn; xuxingguo@mail.dlut.edu.cn; zxsu@dlut.edu.cn).

Yuhao Wang is with the School of Future Technology, School of Artificial Intelligence, Dalian University of Technology, Dalian, 116024, China (e-mail: 924973292@mail.dlut.edu.cn).

Cong Wang and Yang Yang are with the Department of Radiology and Biomedical Imaging, University of California, San Francisco, 94107, USA (e-mail: supercong94@gmail.com; yang.yang4@ucsf.edu).

Jinshan Pan is with the School of Computer Science and Engineering, Nanjing University of Science and Technology, Nanjing, 210094, China (E-mail: sdluran@gmail.com).

}

}

\markboth{}{}

\maketitle

\begin{abstract}
Multi-modal object Re-Identification (ReID) aims to retrieve target instances by leveraging complementary information across modalities. 
However, existing methods suffer from two challenges. 
First, they often fail to exploit well-aligned and reliable semantic priors, making them vulnerable to background clutter and cross-modal misalignment.
On the other hand, they typically rely on holistic feature modeling, overlooking the synergy between global and local representations. 
To overcome these limitations, we propose a robust multi-modal ReID framework with dual semantic guidance and global-local mutual modulation, which mainly consists of three key components, namely the Text-Semantic Injector (TSI), the Masked Global-Local Modulator (MGLM), and the Hierarchical MoE Fusion (HMF).
The TSI enhances semantic awareness by integrating clean and coherent textual features into visual tokens. 
The MGLM enables part-aware cross-modal interaction through joint guidance from soft masks and global context, improving fine-grained feature alignment. 
Finally, the HMF adaptively aggregates multi-spectral features under local semantic supervision, yielding discriminative and robust representations. 
Extensive experiments on three multi-modal ReID benchmarks demonstrate the effectiveness of the proposed method. 
The code will be made publicly available at \url{https://github.com/zw-absin/DSGM} upon acceptance.
\end{abstract}


\begin{figure}
\centering
\includegraphics[width=\linewidth]{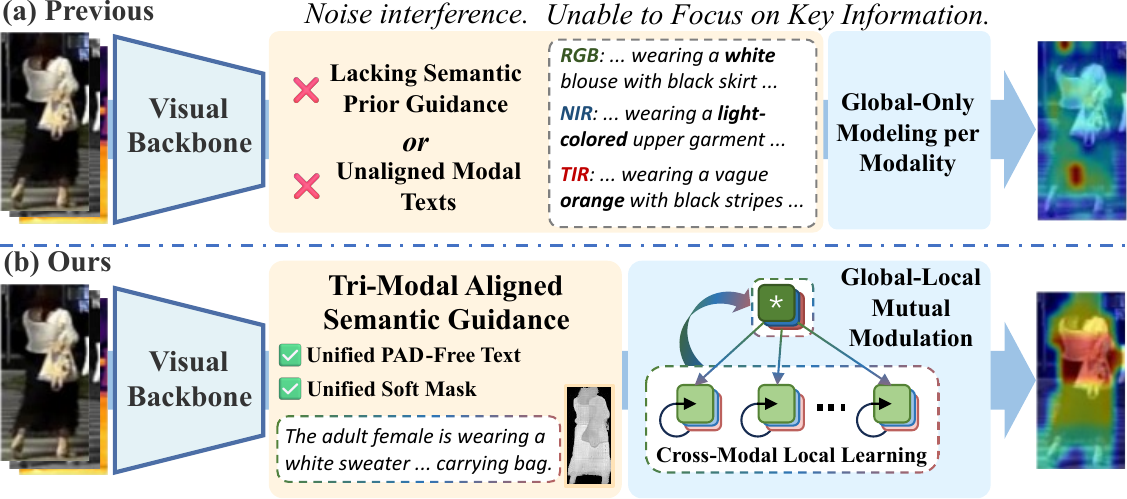}
\caption{(a) Prior approaches that leverage large models often suffer from noisy text features, while other multi-modal methods are typically limited to global modeling.
(b) Our method exploits MLLM-generated textual priors and soft masks to guide attention on key regions and suppress background noise. 
Through a global-local mutual modulation strategy, cross-modal part-aware modeling integrates high-level semantics with structural details, yielding discriminative representations.
}

\label{fig1}
\end{figure}

\section{Introduction}\label{sec:Introduction}

\IEEEPARstart{O}{bject} Re-Identification (ReID) aims to retrieve target instances across non-overlapping camera views, playing a pivotal role in intelligent video surveillance and multi-camera tracking systems. 
Driven by its practical significance, single-modal ReID has achieved remarkable progress in recent years, particularly in the RGB domain~\cite{liu2021watching,liu2023deeply,wang2024other,liu2024video,yu2024tf,ran2025context}.
However, real-world deployment is severely hindered by complex imaging conditions~\cite{wang2020joint,cui2022semi,wang2024perceplie,wang2024correlation,wang2026neural}, such as extreme illumination, adverse weather, heavy occlusion, and dramatic viewpoint variations. These factors degrade the reliability of appearance-based features and limit the robustness of existing approaches.

To mitigate these challenges, multi-modal object ReID has emerged as a promising paradigm that leverages complementary cues from heterogeneous modalities, including visible (RGB), near infrared (NIR), and thermal infrared (TIR) spectra~\cite{wang2024top,pang2024inter,yu2024representation,wu2025enhancing,huang2025harmonizing}. 
Although recent approaches emphasize modality fusion and salient feature selection~\cite{wang2025idea,li2025next}, they often fall short in establishing fine-grained cross-modal alignment owing to insufficient semantic guidance and inadequate modeling of local part-level correspondences. 
As a result, foreground objects cannot be reliably disentangled from cluttered or inconsistent backgrounds, which compromises feature discriminability and yields misaligned representations~\cite{zhang2024magic,ye2021deep}.

A promising direction to address these limitations is the injection of external semantic priors, which has shown significant potential in enhancing ReID robustness. 
Early attempts leverage the aligned image-text embedding space of CLIP~\cite{radford2021learning} to inject semantic priors~\cite{li2023clip,li2025icpl}. However, such priors are typically generic and constrained to pre-defined visual concepts, lacking the specificity required for fine-grained ReID tasks. 
To address this limitation, recent works explore Multi-Modal Large Language Models (MLLMs) for their superior reasoning and cross-modal generalization capabilities~\cite{wang2025idea,li2025next}. 
These methods typically encode textual sequences and either concatenate them with visual tokens through an inverse network~\cite{baldrati2023zero} or use text to directly select salient visual tokens. 
However, such designs lack sufficient high-dimensional interaction between textual and visual representations, limiting the depth of vision-language alignment.
Moreover, existing MLLM-based frameworks commonly adopt modality-specific prompts with fixed padding strategies, as illustrated in Fig.~\ref{fig1}(a). 
This leads to redundant textual encodings and inconsistent semantic descriptions across modalities. 
For instance, a thermal infrared (TIR) image may be described using color terms that do not correspond to its actual grayscale nature, resulting in semantically misleading priors. 
Consequently, these inconsistencies induce misaligned features across modalities, degrading cross-modal representation learning.
Furthermore, to achieve unified cross-modal modeling while preserving modality-specific characteristics, Mixture-of-Experts (MoE) has been widely adopted to model multi-modal interactions~\cite{wang2025decoupled,li2025next}. 
Despite enhanced inter-modality interaction, they predominantly operate on global features and rely on concatenation-based fusion or static routing strategies~\cite{wan2025ugg}.
Such designs overlook fine-grained, part-level semantic correspondences. 
Consequently, they risk misalignment of local regions and fail to leverage contextual cues for adaptive fusion.

In light of these limitations, we argue that robust multi-modal ReID demands structured semantic guidance and hierarchical cross-modal alignment that jointly operate at both global and local levels. 
To this end, we propose a novel framework that unifies semantic priors with dynamic feature modulation to achieve discriminative and robust multi-spectral ReID, as depicted in Fig.~\ref{fig1}(b).
Our approach is built upon two unified semantic signals, namely task-specific textual descriptions and soft semantic masks. 
It further introduces three core components: the Text-Semantic Injector (TSI), the Masked Global-Local Modulator (MGLM), and the Hierarchical MoE Fusion (HMF).
First, the TSI eliminates modality-specific prompt biases by generating unified textual descriptions via a shared MLLM prompting strategy. These descriptions are fused with visual tokens through a hypergraph network, enabling high-order vision-language interaction. To avoid padding-induced noise, irrelevant tokens, such as PAD tokens, are explicitly masked before Generalized Mean (GeM) pooling. This yields clean global semantic embeddings for cross-modal guidance.
Second, the MGLM leverages soft semantic masks, extracted from a pre-trained universal mask generator, to perform part-aware cross-modal modeling. Guided jointly by these masks and global context, MGLM suppresses background clutter and enforces fine-grained alignment of local semantic regions across modalities.
Finally, the HMF employs a two-stage MoE routing mechanism that first selects experts based on local semantic cues and then refines aggregation under global consistency constraints. This hierarchical design enables adaptive fusion that balances local specialization with cross-modal coherence.
The main contributions are summarized as follows:
\begin{itemize}
    \item We present a novel multi-modal ReID framework that integrates dual semantic guidance with global-local mutual modulation, enabling robust cross-modal representation
    learning under challenging conditions.
    
    \item We propose the Text-Semantic Injector (TSI), which uses MLLM-generated textual priors to enable high-order vision-language interaction and enhance global semantic awareness while avoiding padding artifacts.
    
    \item We design the Masked Global-Local Modulator (MGLM) to exploit soft masks and global context for part-aware cross-modal modeling, suppressing background clutter and enhancing local feature alignment.
    
    \item We present the Hierarchical MoE Fusion (HMF), a two-stage routing mechanism that adaptively aggregates multi-modal features under local semantic supervision, balancing specialization and consistency.
\end{itemize}

\begin{figure*}
\centering
\includegraphics[width=\linewidth]{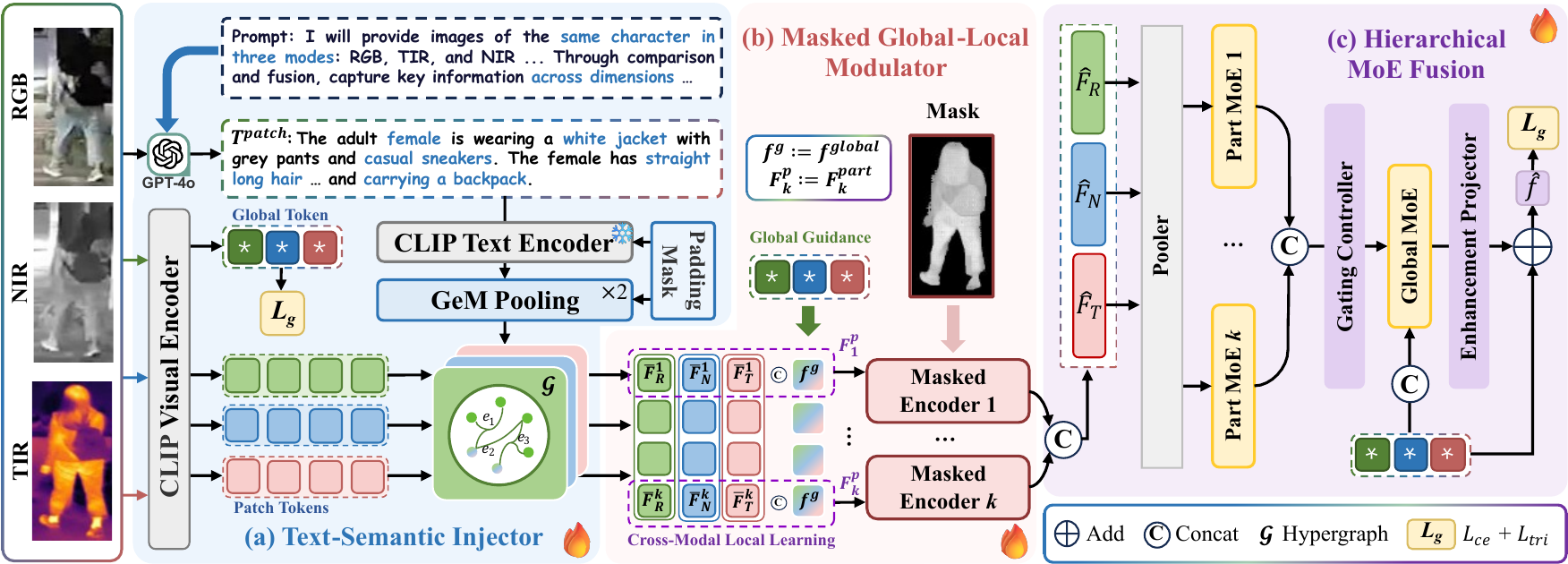}
\caption{Overview of the proposed framework, which employs three modules for fine-grained semantic guidance and part-aware cross-modal feature learning: (a) Text-Semantic Injector (TSI) enables high-order vision-language interaction by fusing global text features with visual tokens via a hypergraph network; (b) Masked Global-Local Modulator (MGLM) enhances local feature learning in key body regions through mask-guided modeling under global context modulation; (c) Hierarchical MoE Fusion (HMF) achieves adaptive trimodal feature aggregation via a multi-level routing mechanism.}
\label{fig:framework}
\end{figure*}

\section{Related Works}\label{sec:RelatedWorks}
\subsection{Multi-Modal Object Re-Identification}
Multi-modal object ReID has gained increasing attention due to its superior robustness in challenging environments by fusing complementary information from multiple sensing modalities~\cite{zheng2021robust, wang2022interact, li2020multi, he2023graph, pan2023progressively, wan2025ugg}. Recent advances leverage the representational power of Vision Transformers (ViT)~\cite{dosovitskiy2020image} to model cross-modal interactions, leading to a series of effective architectures~\cite{pan2023progressively, crawford2023unicat, zhang2024magic, wang2024top}.
Several methods focus on enhancing inter-modal alignment. For example, Wang et al.~\cite{wang2024heterogeneous} propose a test-time training strategy to model relationships among heterogeneous modalities, while Wang et al.~\cite{wang2024top} introduce a cyclic token permutation framework to facilitate cross-modal information exchange. Additionally, CLIP~\cite{radford2021learning} has been adopted as a powerful feature extractor to improve cross-modal alignment~\cite{wang2025mambapro,wang2025decoupled}. Some works employ modality-specific prompts~\cite{wang2025mambapro} or dynamic attention mechanisms~\cite{wang2025decoupled} to better capture modality-specific characteristics.
Despite these advances, most existing methods perform global feature modeling without explicit spatial or semantic decomposition, making them susceptible to background clutter and noise. Moreover, they often lack fine-grained control over local region alignment, limiting their discriminative power in complex scenes.
To overcome these challenges, we introduce a global-local mutual modulation mechanism that leverages dual semantic priors to guide part-aware cross-modal interaction, 
enabling local regions to adaptively align while being informed by the global contextual structure.

\subsection{Semantic-Guided Re-Identification}
Semantic guidance in object ReID has evolved to leverage complementary spatial and linguistic priors for enhanced feature representation. 
These approaches can be broadly categorized into mask-based and text-based methods, which provide structured localization cues and high-level semantic descriptions, respectively.

Mask-based methods utilize segmentation or body parsing models to generate spatial priors for object localization and background suppression. 
Early works employ binary masks to distinguish foreground pedestrians from cluttered backgrounds~\cite{song2018mask, qi2019mask,pang2024inter}. 
To enable finer-grained modeling, body parsing and pose estimation have been further integrated to provide detailed region partitioning or skeletal keypoints~\cite{zhu2020identity, cui2024profd, somers2024keypoint,geng2025pose}. 
For instance, Cui~et al.~\cite{cui2024profd} generate part-aware masks from skeletal keypoints to align structural and semantic features, while Somers~et al.~\cite{somers2024keypoint} perform patch-wise binary classification to localize key body regions. 
However, these methods rely on appearance-dependent structural priors, such as pixel-level segmentation masks or detected keypoints, which are sensitive to visual quality.
Under degraded imaging conditions (e.g., low light, motion blur, or thermal distortion), such cues become unreliable or misaligned across modalities, limiting their robustness in heterogeneous scenarios.

Concurrently, the emergence of large pre-trained models, particularly CLIP~\cite{radford2021learning}, has enabled text-based semantic guidance in ReID by bridging visual and linguistic modalities. 
Learnable textual prompts have been widely adopted to extract aligned image-text representations~\cite{li2023clip, li2025icpl, wei2025multiple}. 
While effective, these prompts are typically optimized in isolation and lack rich, descriptive semantics, constraining their expressiveness in capturing fine-grained attributes.
More recently, Multi-Modal Large Language Models (MLLMs) have been explored to generate more informative and natural language descriptions for ReID~\cite{zhai2024multi, wang2025idea, li2025next}. 
Yet, a critical challenge arises in generating consistent descriptions across heterogeneous modalities: due to significant appearance variations, such as color distortion in infrared images or structural degradation in low-light RGB inputs, MLLMs often produce factually inconsistent narratives.
For example, the same individual may be described with different clothing colors or textures when processed through different modalities, leading to semantic misalignment. 
This inconsistency undermines the reliability of textual guidance for cross-modal representation learning.
As a representative effort, Wang~et al.~\cite{wang2025idea} propose an inverted semantic guidance framework that transforms MLLM-generated text into control signals to steer feature extraction and deformable aggregation. 
However, this approach does not account for the interference introduced by padding tokens during text encoding, which can distort the control signals and adversely affect visual feature modulation.

To address these limitations, we propose a unified multi-modal semantic prior generation pipeline that ensures consistent textual description generation across modalities, enabling fine-grained, robust semantic guidance for multi-modal ReID.

\section{Methodology}
\label{sec:methodology}
\subsection{Overview}
We present a novel framework for multi-modal object Re-Identification (ReID) that leverages dual semantic guidance to enrich feature representations and enable global-local mutual modulation for robust cross-modal learning. 
This design allows the model to focus on critical foreground regions under degraded imaging conditions while jointly capturing global context and fine-grained local details for discriminative representation. 
As illustrated in Fig.~\ref{fig:framework}, the framework comprises three core components: the Text-Semantic Injector (TSI), the Masked Global-Local Modulator (MGLM), and the Hierarchical MoE Fusion (HMF). 
The principle of dual semantic guidance is instantiated through unified textual descriptions and structure-aware soft masks, while global-local mutual modulation is realized by cross-level contextual interaction during both local modeling and global aggregation.

\begin{figure}
\centering
\includegraphics[width=\linewidth]{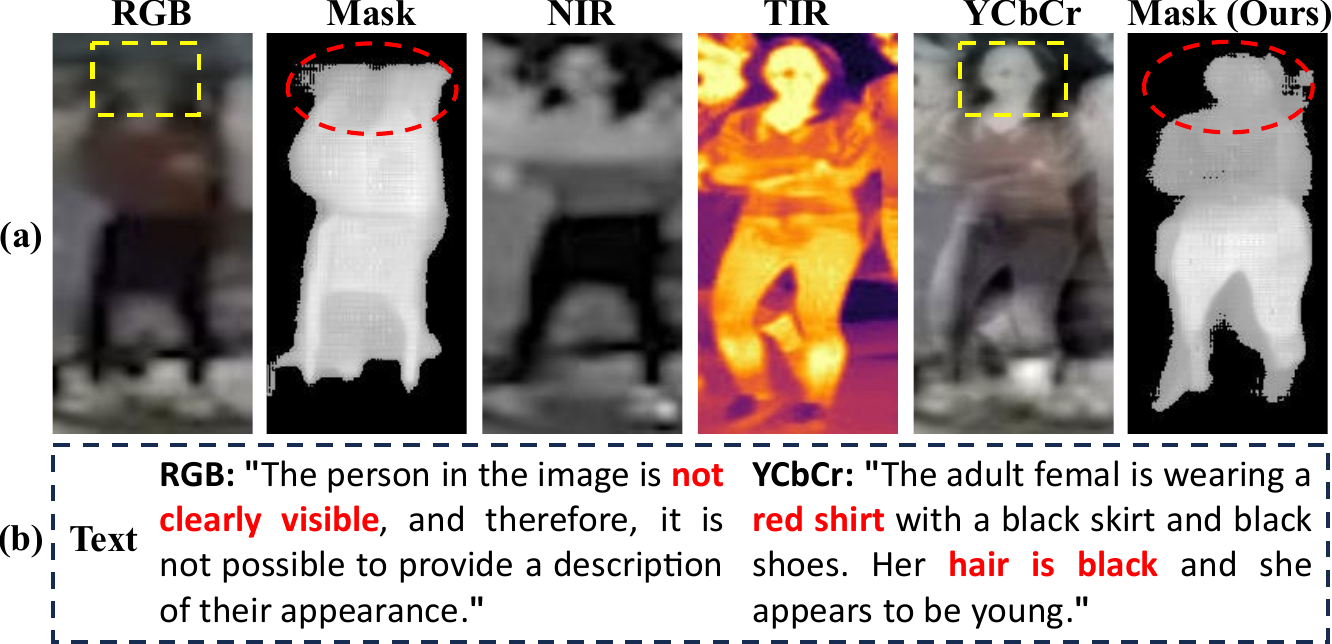}
\caption{(a) Fused image in YCbCr space: the luminance channel (Y) is enhanced by fusing RGB, NIR, and TIR to improve structural details under degraded conditions, while the original chrominance (Cb, Cr) is preserved for color consistency. 
This color space is widely adopted in video standards and enables effective decoupling of structure and color for modality-invariant enhancement. 
(b) MLLM-generated descriptions show improved accuracy and detail when conditioned on the fused image, validating the benefit of multi-modal enhancement for semantic reasoning.
}
\label{fig:mask_y}
\end{figure}

To establish consistent and informative semantic priors, we first exploit the complementarity among RGB, near infrared (NIR), and thermal infrared (TIR) modalities. 
Specifically, we convert the RGB image to the YCbCr color space following~\cite{ITU-RBT601} and extract its luminance channel $I_{\text{RGB}}^Y$, which encodes primary structural information. 
The chrominance channels $C_b$ and $C_r$ preserve color appearance and are retained without modification. 
We then fuse $I_{\text{RGB}}^Y$ with the NIR and TIR images to enhance structural cues under challenging conditions:
\begin{equation}
    I_{\text{fused}}^Y = \alpha \cdot I_{\text{RGB}}^Y + \beta \cdot I_{\text{NIR}} + \gamma \cdot I_{\text{TIR}},
\end{equation}
where $\alpha$, $\beta$, and $\gamma$ are fixed fusion weights. 
The fused luminance channel $I_{\text{fused}}^Y$ is recombined with the original $C_b$ and $C_r$ channels and converted back to RGB, yielding a structure-enhanced image as shown in Fig.~\ref{fig:mask_y}(a). 
This strategy preserves the original color fidelity of RGB while incorporating complementary structural information from non-visible spectra.
The enhanced image is passed to SAM2~\cite{ravi2024sam2} to generate a soft semantic mask $M$, which captures foreground likelihood in a continuous manner. 
To align with the vision encoder's token grid, $M$ is downsampled via max-pooling:
\begin{equation}
    M' = \mathcal{P}_{\text{max}}(M) \in \mathbb{R}^{{N_p}},
\end{equation}
where $N_p$ is the number of image patches.
Unlike conventional mask-based methods that impose hard binary constraints during training, our soft mask serves as a structured semantic prior that is dynamically injected into the network to guide feature modulation without disrupting gradient flow.

In parallel, we adopt the unified prompting strategy from recent works~\cite{wang2025idea,li2025next} to generate a single textual description shared across all modalities.
This description is encoded through one shared text encoder, eliminating the need for multiple modality-specific text branches as in Wang et al.~\cite{wang2025idea}. 
Consequently, our design reduces computational overhead while ensuring consistent semantic grounding across spectra. 
Finally, the original multi-spectral inputs $I_m$ ($m \in \{R, N, T\}$, denoting RGB, NIR, and TIR, respectively) are encoded by a shared CLIP visual encoder, producing patch tokens $F_m$ and a global token $f_m$. 
These visual features, together with the unified textual embedding and the soft semantic mask $M'$, form the dual semantic foundation for subsequent cross-modal refinement in TSI, MGLM, and HMF.

\subsection{Text-Semantic Injector}
\label{subsec:tsi}
To effectively leverage rich textual descriptions generated by MLLMs for multi-modal ReID, we propose the Text-Semantic Injector (TSI), which enables interference-free text encoding and high-order vision-language interaction. 
In this module, we use pooling with trainable parameters to distill unperturbed text features and employ a hypergraph network to fuse semantic and image features, enabling the model to perceive semantic information.

Unlike prior works that ignore padding artifacts in text sequences~\cite{wang2025idea}, TSI introduces a padding mask $M_{\text{pad}} \in \{0,1\}^{N_t}$ {derived from token indices, where the $j$-th element $M_{\text{pad}}^{(j)} = 1$ if the $j$-th token is a padding token and 0 otherwise. This mask is injected into the frozen CLIP text encoder's multi-head self-attention layers to suppress attention at padded positions, ensuring that padding tokens neither attend to nor are attended by valid tokens:}
\begin{equation}
    T^{\text{patch}} = \text{CLIP}_{\text{text}}(T, M_{\text{pad}}) \in \mathbb{R}^{N_t \times D},
\end{equation}
{which yields clean token-level representations. Padded positions are masked out at every self-attention layer, preventing padding artifacts from propagating through the text encoder.}

To further extract rich global semantics, we apply $l$ instances of Generalized Mean (GeM) pooling~\cite{radenovic2018fine} with independently learned exponents. 
Each GeM instance captures a distinct level of semantic granularity: small exponents emphasize fine details, while large exponents focus on dominant concepts.
A validity mask $M_{\text{valid}} = 1 - M_{\text{pad}}$ excludes padded tokens:
\begin{equation}
    T^{\text{global}}_i = \mathcal{P}_{\text{GeM}}^{i}(T^{\text{patch}} \odot M_{\text{valid}}), \quad i \in \{1, \dots, l\}.
\end{equation}
The results are concatenated into a multi-granularity global text vector:
\begin{equation}
    T^{\text{global}} = \left[ T^{\text{global}}_1, \dots, T^{\text{global}}_l \right] \in \mathbb{R}^{l \times D}.
\end{equation}

For fine-grained cross-modal alignment, we employ a Hypergraph-based Feature Refinement (HFR) module $\mathcal{G}^{\text{hyper}}_{m}$, which supports high-order interactions beyond pairwise modeling. 
Hypergraphs naturally encode multi-node correlations~\cite{feng2019hypergraph,sun2025smartfreeedit}, making them ideal for capturing complex semantic structures.
We concatenate and normalize image tokens $F_m$ and $T^{\text{global}}$:
\begin{equation}
    X^{\text{cat}} = \mathrm{LN}\left( \left[ F_m, T^{\text{global}} \right] \right) \in \mathbb{R}^{({N_p} + l) \times D},
\end{equation}
{with $N_p$ being the number of patch tokens.} A dynamic hypergraph is then built from feature similarity, with sparsity controlled by a threshold $\tau$. Message passing yields refined features:
\begin{equation}
    X_m = \mathcal{G}^{\text{hyper}}_{m}(X^{\text{cat}}{,\tau}),
\end{equation}
from which enhanced image tokens are extracted:
\begin{equation}
    \overline{F}_m = X_m^{1:{N_p}} \in \mathbb{R}^{{N_p} \times D}.
\end{equation}
By co-locating image and text tokens in a shared hypergraph space, TSI enables direct, high-order cross-modal interaction. 
Unlike prompt-tuning methods~\cite{li2025icpl} that solely rely on CLIP's pre-trained alignment, TSI leverages MLLM-generated semantics to enrich visual features, thereby enhancing both discriminability and interpretability.
The complete algorithmic procedure of $\mathcal{G}^{\text{hyper}}_{m}$ is detailed in Algorithm~\ref{alg:hypconv}.

\begin{algorithm}[t]
\caption{Hypergraph-based Feature Refinement Module}
\label{alg:hypconv}
\begin{algorithmic}[1]
\Require Features $\mathbf{X} \in \mathbb{R}^{(N+l) \times D}$, threshold $\tau$
\Ensure Refined features $\mathbf{Y} \in \mathbb{R}^{(N+l) \times D}$

\State Compute similarity matrix $\mathbf{S}$ \Comment{e.g., Euclidean}
\State Build adjacency: $\mathbf{H} = \mathbb{I}(\mathbf{S} < \tau)$ \Comment{Binary graph}
\State Project: $\tilde{\mathbf{X}} \gets \text{Linear}(\mathbf{X})$
\State Vertex-to-edge: $\mathbf{E}_i \gets \text{Mean}_{j: H_{ji}=1} (\tilde{\mathbf{X}}_j)$
\State Edge-to-vertex: $\hat{\mathbf{X}}_i \gets \text{Mean}_{j: H_{ij}=1} (\mathbf{E}_j)$
\State Output: $\mathbf{Y} \gets \text{GELU}(\text{LN}(\hat{\mathbf{X}} + \mathbf{X}))$

\State \Return $\mathbf{Y}$
\end{algorithmic}
\end{algorithm}

\subsection{Masked Global-Local Modulator}
\label{subsec:mglm}
Existing multi-modal ReID approaches typically forgo explicit cross-modal part-level modeling. 
This design choice stems from the assumption that powerful architectures such as  CLIP-based variants can implicitly learn global cross-modal correspondences through self-attention or prompt adaptation. 
However, under significant inter-modality appearance discrepancies, such as the absence of  color in thermal imagery or structural degradation in low-light NIR, this implicit modeling may fail to enforce consistent local alignment, leading to suboptimal feature discrimination.
To address this limitation, we argue that explicit cross-modal part correspondence is important for robust multi-modal ReID.
Importantly, this can be achieved with minimal parameter overhead by leveraging semantic priors to guide local feature interaction. 

To achieve robust cross-modal local modeling under degraded conditions, we design the Masked Global-Local Modulator (MGLM), which performs mask-guided and global-aware feature learning to enable aligned part-level representations across modalities.
Given image $I_m$ and mask $M$, we partition them into $k$ horizontal strips, yielding local features $\overline{F}_m^i \in \mathbb{R}^{N_i \times D}$ and binary mask regions $M'_i$ for $i \in \{1, \dots, k\}$, where $N_i = {N_p}/k$. 
This rigid partitioning is effective in practice: boundary refinement strategies tend to be complex, sensitive to occlusion and low resolution, and prone to inducing cross-modal misalignment due to conflicting or unreliable cues across modalities.

To inject global context, we augment each local block with a fused global token $f^{\text{global}} \in \mathbb{R}^D$, which is derived by mapping the trimodal global features $\{f_R, f_N, f_T\}$.
A scalar 1 is prepended to $M'_i$ for dimension alignment. 
The augmented feature and mask for region $i$ are:
\begin{equation}
    F^{\text{part}}_i = \left[ f^{\text{global}},\, \overline{F}_R^i,\, \overline{F}_N^i,\, \overline{F}_T^i \right],
\end{equation}
\begin{equation}
    M^{\text{part}}_i = \left[ 1,\, M'_i,\, M'_i,\, M'_i \right].
\end{equation}
$F^{\text{part}}_i$ is then processed by a mask-aware encoder $\Phi_i$, which performs cross-modal alignment guided by spatial priors:
\begin{equation}
    \tilde{F}_i = \Phi_i \left( F^{\text{part}}_i \odot M^{\text{part}}_i \right),
\end{equation}
where $\odot$ denotes element-wise masking, and $\Phi_i$ consists of MHSA and FFN layers~\cite{vaswani2017attention}.

After local modeling, we decompose $\tilde{F}_i$ (excluding the first global token) into three modality-specific feature blocks:
\begin{equation}
    [\tilde{F}_{i,R},\, \tilde{F}_{i,N},\, \tilde{F}_{i,T}] = \mathcal{C}(\tilde{F}_i^{2:{3N_i+1}}),
\end{equation}
where $\mathcal{C}(\cdot)$ splits $\tilde{F}_i^{2:{3N_i+1}} \in \mathbb{R}^{3N_i \times D}$ into three consecutive segments corresponding to RGB, NIR, and TIR modalities.
Finally, the refined local features from all $k$ regions are concatenated along the token dimension to form the enhanced modality-specific representation:
\begin{equation}
    \hat{F}_m = \left[ \tilde{F}_{1,m},\, \tilde{F}_{2,m},\, \dots,\, \tilde{F}_{k,m} \right], \quad m \in \{R, N, T\}.
\end{equation}
MGLM thus achieves mask-guided, globally modulated local modeling, promoting robust and aligned cross-modal feature learning under challenging conditions. The detailed procedure is summarized in Algorithm~\ref{alg:mglm}.

\begin{algorithm}[t]
\caption{Masked Global-Local Modulator (MGLM)}
\label{alg:mglm}
\begin{algorithmic}[1]
\Require 
    Image features $\{\overline{F}_m^i\}_{m,i}$, masks $\{M'_i\}_{i=1}^k$, global representations $\{f_m\}_{m=R,N,T}$
\Ensure 
    Enhanced features $\{\hat{F}_m\}_{m=R,N,T}$

\State $f^{\text{global}} \gets \text{Fuse}(\{f_m\})$ \Comment{Unified global context}

\For{$i = 1$ to $k$}
    \State $F^{\text{part}}_i \gets \left[ f^{\text{global}},\, \overline{F}_R^i,\, \overline{F}_N^i,\, \overline{F}_T^i \right]$ \Comment{Augment with global}
    \State $M^{\text{part}}_i \gets \left[ 1,\, M'_i,\, M'_i,\, M'_i \right]$ \Comment{Mask alignment}
    \State $\tilde{F}_i \gets \Phi_i \left( F^{\text{part}}_i \odot M^{\text{part}}_i \right)$ \Comment{Mask-guided encoding}
    \State $[\tilde{F}_{i,R}, \tilde{F}_{i,N}, \tilde{F}_{i,T}] \gets \mathcal{C}(\tilde{F}_i^{2:3N_i+1})$ \Comment{Split per modality}
\EndFor

\State $\hat{F}_m \gets \left[ \tilde{F}_{1,m},\, \dots,\, \tilde{F}_{k,m} \right],\ \forall m \in \{R,N,T\}$ 

\State \Return $\{\hat{F}_m\}$
\end{algorithmic}
\end{algorithm}

\begin{algorithm}[t]
\caption{Context-Aware Mixture-of-Experts (MoE)}
\label{alg:moe}
\begin{algorithmic}[1]
\Require Input $x \in \mathbb{R}^{3D}$, context $c \in \mathbb{R}^{D}$ (optional)
\Ensure Output $o \in \mathbb{R}^{D}$, auxiliary loss $\ell$

\State $z \gets W_g x$ where $W_g \in \mathbb{R}^{E \times 3D}$ \Comment{Compute gating logits}

\State $\Delta z \gets \begin{cases}
    W_c c & \text{if } c \text{ is provided} \\
    \mathbf{0} & \text{otherwise}
\end{cases}$ \Comment{Context bias}

\State $\tilde{z} \gets z + \Delta z$ \Comment{Local-to-global modulation}

\State $\tilde{z} \gets \tilde{z} + \epsilon \odot \mathrm{softplus}(W_n x)$ \Comment{Add noise}

\State $\mathcal{S} = \mathrm{top}_K(\tilde{z})$ \Comment{Select top-$K$ experts}

\State $g_i = \mathrm{softmax}(\tilde{z})_{e_i}$ for $e_i \in \mathcal{S}$ \Comment{Compute weights}

\State $o_i = \mathcal{E}_{e_i}(x)$ \Comment{Forward selected experts}

\State $o = \sum_{i=1}^K g_i \cdot o_i$ \Comment{Aggregate}

\State $\ell = \lambda \sum_{j=1}^E \rho_j P_j$ \Comment{Compute load-balancing loss}

\State \Return $o, \ell$
\end{algorithmic}
\end{algorithm}

\subsection{Hierarchical MoE Fusion}
\label{subsec:hierarchical_moe}

We propose a Hierarchical MoE Fusion (HMF) framework that integrates local part-level and global holistic representations through a two-stage routing mechanism.
Our HMF comprises two components: (1) a Part-Aware MoE for discriminative local patterns, and (2) a Contextual Global MoE that leverages part-level semantics to guide the fusion of global features, enabling adaptive and context-sensitive integration.

We first define a context-aware MoE as the building block for both stages. 
Then, based on this building block, we further propose Part-Aware MoE and Contextual Global MoE.
Given input $x \in \mathbb{R}^{3D}$ formed by concatenating multi-modal features, a gating network routes $x$ to $K=2$ of $E=4$ experts. The gating logits $z = W_g x$ are modulated by a context signal $c \in \mathbb{R}^D$, yielding the context-aware logits:
\begin{equation}
    \tilde{z} = z + W_c c,
\end{equation}
enabling context-aware routing. 
During training, we adopt noisy gating~\cite{shazeer2017outrageously} to encourage expert exploration:
\begin{equation}
    \tilde{z} \gets \tilde{z} + \epsilon \odot \mathrm{softplus}(W_n x),
\end{equation}
{where $W_n \in \mathbb{R}^{E \times 3D}$ projects $x$ to per-expert noise scales. $\epsilon \in \mathbb{R}^{E}$ is a standard Gaussian noise vector with independent entries $\epsilon_j \sim \mathcal{N}(0,1)$, and $\odot$ denotes element-wise multiplication. Noisy gating with $\mathrm{softplus}(W_n x)$ is applied only during training.}
The output is computed as a weighted sum of the activated experts:
\begin{equation}
    o = \sum_{i=1}^K g_i \cdot \mathcal{E}_{e_i}(x), \quad g_i = \mathrm{softmax}(\tilde{z}_{\mathcal{S}})_i,
\end{equation}
{where $\mathcal{S}=\mathrm{top}_K(\tilde{z})$ collects the indices of the top-$K$ experts. $e_i \in \mathcal{S}$ is the $i$-th selected expert index, $\mathcal{E}_{e_i}(\cdot)$ is the corresponding two-layer MLP expert, and $g_i$ is normalized by softmax over $\tilde{z}_{\mathcal{S}}=[\tilde{z}_{e_1},\ldots,\tilde{z}_{e_K}]$ only.}
An auxiliary load-balancing loss $\mathcal{L}_{\text{aux}} = \lambda \sum_{j=1}^E \rho_j P_j$ ensures expert diversity, 
where $\lambda$ controls the strength of the balancing regularization.
{Here $\rho_j$ is the dispatch frequency of expert $j$ in a mini-batch. $P_j$ is its batch-averaged router probability over all $E$ logits. $\mathcal{L}_{\text{aux}}$ is computed during training to promote balanced expert utilization and avoid router collapse.} 
Experts are grouped to enable efficient batched execution, 
and the overall MoE procedure is detailed in Algorithm~\ref{alg:moe}.

\subsubsection{Part-Aware MoE}
We partition each modality into $k$ horizontal strips and apply adaptive pooling to obtain part-level features $f^i_m = \mathcal{P}_{\text{adap}}(\hat{F}_m^i)$. For each part $i$, we form $x_i = [f^i_R, f^i_N, f^i_T]$ and process it with Part-Aware MoE:
\begin{equation}
    f'_i,\, \mathcal{L}_{\text{part}}^i = \mathcal{M}_{\text{part}}(x_i).
\end{equation}
The total part loss and context vector are:
\begin{equation}
    \mathcal{L}_{\text{part}} = \sum_i \mathcal{L}_{\text{part}}^i, \quad
    c_{\text{part}} = \frac{1}{k} \sum_i f'_i,
\end{equation}
where $c_{\text{part}}$ summarizes local semantics for global guidance.

\subsubsection{Contextual Global MoE}
The Global MoE $\mathcal{M}_{\text{global}}$ takes $x_{\text{global}} = [f_R, f_N, f_T]$ and uses $c_{\text{part}}$ as a modulation signal:
\begin{equation}
    f_{\text{global}},\, \mathcal{L}_{\text{global}} = \mathcal{M}_{\text{global}}(x_{\text{global}}, c_{\text{part}}).
\end{equation}
To preserve modality-specific content, we enhance each global feature via a residual connection:
\begin{equation}
    \hat{f}_m = f_m + W_r f_{\text{global}}, \quad m \in \{R,N,T\}.
\end{equation}
The fused features $\hat{f} = [\hat{f}_R, \hat{f}_N, \hat{f}_T]$ are used for classification. Total auxiliary loss: $\mathcal{L}_{\text{aux}} = \mathcal{L}_{\text{part}} + \mathcal{L}_{\text{global}}$.

\subsection{Optimization}
\label{subsec:loss}

To optimize the network, the following loss is used:
\begin{equation}
    \mathcal{L} = \mathcal{L}_g([f^{class}_{R}, f^{class}_{N}, f^{class}_{T}]) + \mathcal{L}_g(\hat{f}) + \mathcal{L}_{\mathrm{aux}},
\end{equation}
where $\mathcal{L}_g(\mathcal{X})$ refers to the label-smoothed cross-entropy and triplet loss~\cite{szegedy2016rethinking,hermans2017defense} on  the features from the visual encoder and HMF:
\begin{equation}
    \mathcal{L}_g(\mathcal{X}) = \mathcal{L}_{\mathrm{ce}}(\mathcal{X}) + \mathcal{L}_{\mathrm{tri}}(\mathcal{X}),
\end{equation}
where $\mathcal{X}$ denotes the input features.

\section{Experiments}\label{sec:HC}

\subsection{Datasets and Evaluation Metrics}
We evaluate our method on three widely used multi-modal ReID benchmarks, covering both person and vehicle scenarios.
To enable scalable and hardware-independent annotation of multi-spectral images, 
automated textual descriptions are generated using the API-based GPT-4o~\cite{hurst2024gpt}, 
while mask generation is performed with SAM2~\cite{ravi2024sam2}.

\textbf{RGBNT201}~\cite{zheng2021robust}: A large-scale person ReID dataset with 201 identities captured under four non-overlapping cameras. Each identity is synchronously recorded in RGB, NIR, and TIR modalities. In total, the dataset contains 4,787 aligned images per modality, divided into 141 identities for training, 30 for validation, and 30 for testing. It includes common real-world challenges such as pose/viewpoint variations, occlusion, cluttered backgrounds, as well as severe degradations from nighttime, strong illumination, haze, and fog.

\textbf{RGBNT100}~\cite{li2020multi}: A vehicle ReID dataset extended from RGBN300 by additionally collecting 17,250 TIR images paired with corresponding RGB–NIR samples. It covers 100 vehicles with 8,675 triplets for training and 8,575 triplets for testing/gallery. The dataset introduces challenges such as viewpoint changes, abnormal lighting, and partial occlusion.

\textbf{MSVR310}~\cite{zheng2023cross}: A smaller yet more challenging vehicle ReID dataset with 2,087 image triplets captured under adverse conditions including heavy occlusion, low illumination, and diverse weather scenarios.

\textbf{Evaluation Metrics.} Following common practice, we report mean Average Precision (mAP) and Cumulative Matching Characteristics (CMC) at Rank-$K$ ($K=1,5,10$).

\begin{table}
    \caption{Performance comparison on RGBNT201. 
    Methods with $\dagger$ are CLIP-based, those with $*$ are ViT-based, and the rest are CNN-based}
  \label{tab:multi-spectral person ReID}
  \centering
  \renewcommand\arraystretch{1.2}
  \setlength{\tabcolsep}{6.35pt}
  \begin{tabular}{cr|cccc}
      \noalign{\hrule height 1pt}
  &\multicolumn{1}{c|}{\multirow{2}{*}{Methods}}   & \multicolumn{4}{c}{RGBNT201} \\ \cline{3-6}
            & & mAP & R-1 & R-5 & R-10 \\ \hline
            \multirow{3}{*}{\rotatebox{90}{Single}}
            &OSNet~\cite{zhou2019omni}~(ICCV19)  & 25.4 & 22.3 & 35.1 & 44.7 \\
            &CAL~\cite{rao2021counterfactual}~(ICCV21)  & 27.6 & 24.3 & 36.5 & 45.7 \\
            &PCB~\cite{sun2018beyond}~(ECCV18)  & 32.8 & 28.1 & 37.4 & 46.9 \\ \hline
            \multirow{20}{*}{\rotatebox{90}{Multi-Modal}}
            & HAMNet~\cite{li2020multi}~(AAAI20)   & 27.7         & 26.3            & 41.5            & 51.7             \\
            & PFNet~\cite{zheng2021robust}~(AAAI21)    & 38.5         & 38.9            & 52.0            & 58.4             \\
            & IEEE~\cite{wang2022interact}~(AAAI22)     & 47.5         & 44.4            & 57.1            & 63.6             \\
            & DENet~\cite{zheng2023dynamic}~(TNNLS23)    & 42.4         & 42.2            & 55.3            & 64.5            \\
            & LRMM~\cite{wu2025lrmm}~(ESWA25) & 52.3 & 53.4 & 64.6 & 73.2\\
            & UniCat$^*$~\cite{crawford2023unicat}~(NIPSW23)   & 57.0         & 55.7            & -            & -            \\
            & HTT$^*$~\cite{wang2024heterogeneous}~(AAAI24) &71.1 &73.4 &83.1 &87.3\\
            & TOP-ReID$^*$~\cite{wang2024top}~(AAAI24)  &72.3 &76.6 &84.7 &89.4\\
            & EDITOR$^*$~\cite{zhang2024magic}~(CVPR24) & 66.5       & 68.3           & 81.1        & 88.2             \\
            & RSCNet$^*$~\cite{yu2024representation}~(TCSVT24) & 68.2 & 72.5 & - & - \\
            & WTSF-ReID$^*$~\cite{yu2025wtsf}~(ESWA25) & 67.9 &72.2 &83.4 &89.7 \\
            & DESANet$^*$~\cite{dong2025escaping}~(TIP25) & 74.6 &77.6 &87.1 &91.3 \\
            & MambaPro$^\dagger$~\cite{wang2025mambapro}~(AAAI25) & 78.9 & 83.4 & {89.8} & 91.9 \\
            & DeMo$^\dagger$~\cite{wang2025decoupled}~(AAAI25)  &{79.0}      &{82.3}      &88.8      &{92.0}      \\
            & IDEA w/o Text$^\dagger$~\cite{wang2025idea}~(CVPR25)  &74.5      &75.0      &84.8      &88.8     \\
            & IDEA$^\dagger$~\cite{wang2025idea}~(CVPR25)  &80.2      &82.1      &90.0      &93.3      \\
            & MDReID$^\dagger$~\cite{feng2025mdreid}~(NeurIPS25) & 82.1 & 85.2 & 90.3 & 92.6 \\
            & NEXT$^\dagger$~\cite{li2025next}~(arXiv25) & 82.4 & 86.6 & 92.0 & 94.7 \\
            & Signal$^\dagger$~\cite{liu2026signal}~(AAAI26) & 80.3 & 85.2 & 91.4 & 93.7 \\
            & MGRNet$^\dagger$~\cite{wan2026mgrnet}~(TIFS26) & 80.5 & 85.0 & 90.0 & 92.6 \\
            \rowcolor[gray]{0.92}
            & $\textbf{Ours}^\dagger$  &82.6      &87.0      &92.0      &93.9      \\
  \noalign{\hrule height 1pt}
  \end{tabular}
\end{table}

\subsection{Implementation Details}
Our framework is implemented using PyTorch on an NVIDIA A800 GPU, with CLIP~\cite{radford2021learning} serving as the visual backbone. Input images are resized to $256 \times 128$ for RGBNT201 and to $128 \times 256$ for both RGBNT100 and MSVR310. We apply standard data augmentation techniques, including random horizontal flipping, random cropping, and random erasing~\cite{zhong2020random}, to improve model robustness.
During training, we use a mini-batch size of 64 with 8 images sampled per identity for RGBNT201 and MSVR310, and a mini-batch size of 128 with 16 images per identity for RGBNT100. The total number of training epochs is set to 60 for RGBNT201 and to 50 for RGBNT100 and MSVR310. Our modules are fine-tuned with an initial learning rate of $3.5 \times 10^{-6}$, which is decayed to $3.5 \times 10^{-7}$ over the course of training.
The text generation procedure, as well as the training and inference protocols, follow the settings established in prior works~\cite{wang2025idea,li2025next}. Following common practice in Mixture-of-Experts (MoE) training, we set the auxiliary loss weight $\lambda = 10^{-2}$.

\subsection{Comparison with State-of-the-Art Methods}

\subsubsection{Multi-Modal Person ReID}
Table~\ref{tab:multi-spectral person ReID} compares our method with state-of-the-art (SOTA) multi-modal approaches on the RGBNT201 dataset. 
Our approach consistently outperforms LRMM~\cite{wu2025lrmm}, demonstrating a significant advantage over the most recent CNN-based method on multi-modal data. 
By integrating dual semantic guidance and global-local mutual modulation, Ours$^\dagger$ achieves 82.6\% mAP and 87.0\% Rank-1 accuracy, surpassing TOP-ReID$^*$~\cite{wang2024top} by 10.3\% and 10.4\%, respectively. 
Moreover, our method exceeds CLIP-based approaches such as DeMo$^\dagger$~\cite{wang2025decoupled} and IDEA$^\dagger$~\cite{wang2025idea} across all evaluation metrics (mAP, Rank-1, Rank-5, Rank-10). 
Notably, it improves upon IDEA$^\dagger$ by 2.4\% in mAP and 4.9\% in Rank-1. 
{Compared with the recent MoE-based method NEXT$^\dagger$~\cite{li2025next}, our method achieves higher mAP and Rank-1 accuracy and the same Rank-5 score, while NEXT$^\dagger$ attains a higher Rank-10.
This difference is attributable to NEXT's parallel semantic and structural expert branches fused via Cross-Attention, which can be advantageous for retrieval metrics that benefit from broad candidate coverage.
In contrast, our HMF adopts a two-stage local-to-global routing mechanism that enables dynamic, context-aware feature aggregation, yielding stronger performance on the key metrics of mAP and Rank-1.}
These results validate that the combination of textual and mask-based semantic guidance with global-local feature modulation substantially enhances feature discriminability in multi-modal person ReID.

\begin{table}
    \caption{Performance comparison on RGBNT100 and MSVR310}
    \centering
    \vspace{-0.2mm}
    \renewcommand\arraystretch{1.15}
    \setlength{\tabcolsep}{6.85pt}
    \begin{tabular}{cr|cccc}
        \noalign{\hrule height 1pt}
    &\multicolumn{1}{c|}{\multirow{2}{*}{Methods}} &  \multicolumn{2}{c}{RGBNT100} & \multicolumn{2}{c}{MSVR310} \\\cline{3-6}
            & & mAP & R-1 & mAP & R-1 \\
            \hline
            \multirow{3}{*}{\rotatebox{90}{Single}}
            &PCB~\cite{sun2018beyond}~(ECCV18)& 57.2 & 83.5 & 23.2 & 42.9 \\
            &OSNet~\cite{zhou2019omni}~(ICCV19)& 75.0 & 95.6 & 28.7 & 44.8 \\
            &TransReID$^*$~\cite{he2021transreid}~(ICCV21)& 75.6 & 92.9 & 18.4 & 29.6 \\
            \hline
            \multirow{23}{*}{\rotatebox{90}{Multi-Modal}}
            &HAMNet~\cite{li2020multi}~(AAAI20) & 74.5 & 93.3 & 27.1 & 42.3 \\
            &PFNet~\cite{zheng2021robust}~(AAAI21)& 68.1 & 94.1 & 23.5 & 37.4 \\
            &GAFNet~\cite{guo2022generative}~(ICSP22) & 74.4 & 93.4 & - & - \\
            &GPFNet~\cite{he2023graph}~(TITS23) & 75.0 & 94.5 & - & - \\
            &CCNet~\cite{zheng2023cross}~(INFFUS23) & 77.2 & 96.3 & 36.4 & 55.2 \\
            & LRMM~\cite{wu2025lrmm}~(ESWA25) & 78.6 & 96.7 & 36.7 &49.7\\
            &GraFT$^*$~\cite{yin2023graft}~(arXiv23) &76.6 &94.3 &- &-\\
            &UniCat$^*$~\cite{crawford2023unicat}~(NIPSW23)    & 79.4         & 96.2  & -            & -            \\
            &PHT$^*$~\cite{pan2023progressively}~(Sensors23) & 79.9 & 92.7 & - & - \\
            & HTT$^*$~\cite{wang2024heterogeneous}~(AAAI24) &75.7&92.6&- &-\\
            & TOP-ReID$^*$~\cite{wang2024top}~(AAAI24) &81.2 & 96.4 & 35.9 & 44.6 \\
            & EDITOR$^*$~\cite{zhang2024magic}~(CVPR24) & 82.1 & 96.4 &39.0 & 49.3\\
            & FACENet$^*$~\cite{zheng2025flare}~(INFFUS25) & 81.5 &{96.9} &36.2 &54.1 \\
            & RSCNet$^*$~\cite{yu2024representation}~(TCSVT24) &82.3 &96.6 &39.5 &49.6\\
            & WTSF-ReID$^*$~\cite{yu2025wtsf}~(ESWA25) & 82.2 &96.5 & 39.2 & 49.1 \\
            & DESANet$^*$~\cite{dong2025escaping}~(TIP25) & 82.1 &97.4 &39.2 &47.8 \\
            & MambaPro$^\dagger$~\cite{wang2025mambapro}~(AAAI25) & 83.9 & 94.7 &{47.0} & 56.5 \\
            & DeMo$^\dagger$~\cite{wang2025decoupled}~(AAAI25) &{86.2}      &97.6 &49.2      &{59.8} \\
            & IDEA$^\dagger$~\cite{wang2025idea}~(CVPR25)& 87.2      &96.5 &{47.0}      &62.4 \\
            & MDReID$^\dagger$~\cite{feng2025mdreid}~(NeurIPS25) & 85.3 & 95.6 & 51.0 & 68.9 \\
            & NEXT$^\dagger$~\cite{li2025next}~(arXiv25) & 88.2 & 97.7 & 60.8 & 79.0 \\
            & Signal$^\dagger$~\cite{liu2026signal}~(AAAI26) & 86.3 & 97.6 & 53.6 & 71.9 \\
            & MGRNet$^\dagger$~\cite{wan2026mgrnet}~(TIFS26) & 88.2 & 98.0 & 53.2 & 67.2 \\
            \rowcolor[gray]{0.92}
            & $\textbf{Ours}^\dagger$&89.4      &98.2 &64.6      &76.0 \\
    \noalign{\hrule height 1pt}
    \end{tabular}
    \label{tab:multi-spectral vehicle ReID}
\end{table}

\begin{table}
\caption{Incremental ablation study of the main modules}
  \centering
  \renewcommand\arraystretch{1.0}
  \setlength{\tabcolsep}{6.85pt}
  \begin{tabular}{ccccccc}
      \noalign{\hrule height 1pt}
      \multicolumn{1}{c}{\multirow{2}{*}{\textbf{Index}}} &\multicolumn{3}{c}{\textbf{Modules}} & \multicolumn{2}{c}{\textbf{Metrics}} &\multicolumn{1}{c}{\multirow{2}{*}{\textbf{FLOPs(G)}}}\\
      \cmidrule(r){2-4} \cmidrule(r){5-6}
 & \textbf{TSI}              & \textbf{MGLM}                & \textbf{HMF}                   & \textbf{mAP}    & \textbf{Rank-1}   \\\hline
  A                  & \XSolidBrush                  & \XSolidBrush                  & \XSolidBrush                    & 70.3  & 72.1 &{34.27}\\
  B                  & \CheckmarkBold                  & \XSolidBrush                  & \XSolidBrush                      & 76.1  & 79.1 &37.45\\
  \multirow{1}{*}{C} & \multirow{1}{*}{\CheckmarkBold} & \multirow{1}{*}{\CheckmarkBold} & \multirow{1}{*}{\XSolidBrush}    & 78.8  & 84.0 &37.89\\
  \rowcolor[gray]{0.92}
  \multirow{1}{*}{D} & \multirow{1}{*}{\CheckmarkBold} & \multirow{1}{*}{\CheckmarkBold} & \multirow{1}{*}{\CheckmarkBold}    &\textbf{82.6} &\textbf{87.0}  &37.91\\
  \noalign{\hrule height 1pt}
  \end{tabular}
  \label{tab:main_ablation}
\end{table}

\begin{table}[t]
\caption{Comparison of training efficiency on the RGBNT201 dataset}
  \centering
  \setlength{\tabcolsep}{20.75pt}
  \begin{tabular}{l|cc}
    \noalign{\hrule height 1pt}
    \textbf{Model} & \textbf{FLOPs(G)} & \textbf{Samples/s}  \\ \hline
    TOP-ReID$^*$~\cite{wang2024top} & 35.51	 & 168.03  \\
    IDEA$^\dagger$~\cite{wang2025idea} & 43.73 & 159.68  \\
    \rowcolor[gray]{0.92}
    \textbf{Ours$^\dagger$} & \textbf{37.91} & \textbf{169.40}  \\
    \noalign{\hrule height 1pt}
  \end{tabular}
  \label{tab:flops}
\end{table}

\begin{table}
    \caption{Comparison of the number of trainable parameters among different methods}
    \centering
    \renewcommand\arraystretch{1.2}
    \setlength{\tabcolsep}{7pt}
    \begin{tabular}{r|c|cccc}
        \noalign{\hrule height 1pt}
    \multicolumn{1}{c|}{\multirow{2}{*}{\textbf{Methods}}} &
    \multicolumn{1}{c|}{\multirow{2}{*}{\textbf{Params (M)}}} & \multicolumn{2}{c}{\textbf{RGBNT201}} & \multicolumn{2}{c}{\textbf{MSVR310}} \\\cline{3-6}
    && \textbf{mAP} & \textbf{R-1} & \textbf{mAP} & \textbf{R-1} \\
    \hline
    TOP-ReID$^*$~\cite{wang2024top} &324.53 &72.3 & 76.6 & 35.9 & 44.6 \\
    EDITOR$^*$~\cite{zhang2024magic} &118.55& 66.5 & 68.3 &39.0 & 49.3\\
    RSCNet$^*$~\cite{yu2024representation} &124.10&68.2 &72.5 &39.5 &49.6\\
    WTSF-ReID$^*$~\cite{yu2025wtsf} &143.60& 67.9 &72.2 & 39.2 & 49.1 \\
    \rowcolor[gray]{0.92}
    $\textbf{Ours}^\dagger$&\textbf{100.35}&\textbf{82.6} 	&\textbf{87.0} &\textbf{64.6}	&\textbf{76.0} \\
    \noalign{\hrule height 1pt}
    \end{tabular}
    \label{tab:params}
\end{table}

\begin{table}[t]
\caption{Ablation of internal design choices for each module}
  \centering
  \setlength{\tabcolsep}{8.25pt}
  \begin{tabular}{c|ll|cc}
    \noalign{\hrule height 1pt}
    \textbf{Index} &\multicolumn{2}{c|}{\textbf{Model Variant}}  & \textbf{mAP} & \textbf{R-1}  \\ \hline
    A &\multicolumn{1}{c|}{\multirow{3}{*}{\textbf{TSI}}}&  w/o Text & 79.4 & 80.5  \\
    B &\multicolumn{1}{c|}{} &w/o GeM Pooling & 80.4 & 82.9  \\
    C &\multicolumn{1}{c|}{} &w/o Padding Mask & 82.1 & 85.0  \\ \hline
    D &\multicolumn{1}{c|}{\multirow{3}{*}{\textbf{MGLM}}} & w/o Mask & 79.5 & 81.2  \\
    E & \multicolumn{1}{c|}{}& w/o Trimodal Collaboration & 79.6 & 81.6  \\
    F &\multicolumn{1}{c|}{} & w/o Global Input & 79.3 & 83.0  \\ \hline
    G &\multicolumn{1}{c|}{\multirow{2}{*}{\textbf{HMF}}}& w/o Global MoE & 78.3 & 80.6  \\
    H & \multicolumn{1}{c|}{}& w/o Part-Aware MoE & 79.1 & 83.3  \\\hline
    \rowcolor[gray]{0.92}
    I &\multicolumn{2}{c|}{\textbf{Ours (Full Model)}}    & \textbf{82.6} & \textbf{87.0}  \\
    \noalign{\hrule height 1pt}
  \end{tabular}
  \label{tab:sub_ablation}
\end{table}

\begin{table*}[t]
\centering
\caption{\textbf{Fine-Grained Analysis} on RGBNT201. 
Performance under different design choices. Default settings: $l=2$, $\tau=2$, TSI w/ HFR, MGLM w/ RGB-only global token, $k=4$, HMF w/ contextual gate modulation. Default configurations are highlighted in gray}
\label{tab:ablations}
\vspace{-5mm}
\begin{tabular}{lll}
\subfloat[
{\footnotesize\selectfont \textbf{GeM Count $l$}. Performance peaks at $l=2$.}
\label{tab:number_gem}
]{
\begin{minipage}{0.35\linewidth}
\centering
\begin{tabular}{x{18}x{24}x{24}}
$l$ & mAP & R-1 \\
\shline
1 & 80.9 & 83.7 \\
\cellcolor{gray!20} 2 & \cellcolor{gray!20} \textbf{82.6} & \cellcolor{gray!20} \textbf{87.0} \\
3 & 80.3 & 83.9 \\
4 & 79.3 & 83.3 \\
\end{tabular}
\end{minipage}
} &
\hspace{-7.5mm}
\subfloat[
{\footnotesize\selectfont \textbf{HFR Threshold $\tau$}. Performance by $\tau$.}
\label{tab:hg_threshold}
]{
\begin{minipage}{0.31\linewidth}
\centering
\begin{tabular}{x{18}x{24}x{24}}
$\tau$ & mAP & R-1 \\
\shline
0.5 & 80.1 & 82.5 \\
1 & 82.0 & 84.6 \\
\cellcolor{gray!20} 2 & \cellcolor{gray!20} \textbf{82.6} & \cellcolor{gray!20} \textbf{87.0} \\
3 & 82.6 & 84.8 \\
4 & 79.7 & 81.9 \\
\end{tabular}
\end{minipage}
} &
\hspace{-7.5mm}
\subfloat[
{\footnotesize\selectfont \textbf{Text Integration}. HFR outperforms attention.}
\label{tab:hfr}
]{
\begin{minipage}{0.33\linewidth}
\centering
\begin{tabular}{y{65}x{24}x{24}}
Case & mAP & R-1 \\
\shline
w/ CrossAttn & 77.9 & 80.7 \\ 
w/ HFR (shared) & 80.4 & 84.0 \\
\cellcolor{gray!20} w/ HFR (sep) & \cellcolor{gray!20} \textbf{82.6} & \cellcolor{gray!20} \textbf{87.0} \\
\end{tabular}
\end{minipage}
} \\
\subfloat[
{\footnotesize\selectfont \textbf{Global Token in MGLM}. RGB-only is best.}
\label{tab:global_MGLM}
]{
\begin{minipage}{0.35\linewidth}
\centering
\begin{tabular}{y{40}x{24}x{24}}
Case & mAP & R-1 \\
\shline
\cellcolor{gray!20} RGB-only & \cellcolor{gray!20} \textbf{82.6} & \cellcolor{gray!20} \textbf{87.0} \\
Sum Fusion & 80.1 & 83.6 \\
Concat & 81.3 & 83.9 \\
\end{tabular}
\end{minipage}
} &
\hspace{-7.5mm}
\subfloat[
{\footnotesize\selectfont \textbf{Partitions $k$}. Horizontal split ablation.}
\label{tab:partitions_k}
]{
\begin{minipage}{0.31\linewidth}
\centering
\begin{tabular}{y{8}x{18}x{18}x{38}}
$k$ & mAP & R-1 & Param (M) \\
\shline
2 & 78.8 & 82.9 & 94.04 \\
\cellcolor{gray!20} 4 & \cellcolor{gray!20} \textbf{82.6} & \cellcolor{gray!20} \textbf{87.0} & \cellcolor{gray!20} \textbf{100.35} \\
8 & 78.1 & 81.7 & 112.96 \\
\end{tabular}
\end{minipage}
} &
\hspace{-7.5mm}
\subfloat[
{\footnotesize\selectfont \textbf{HMF Interaction}. Contextual gate wins.}
\label{tab:hmf_types}
]{
\begin{minipage}{0.33\linewidth}
\centering
\begin{tabular}{y{65}x{24}x{24}}
Case & mAP & R-1 \\
\shline
\cellcolor{gray!20} Gate Modulation & \cellcolor{gray!20} \textbf{82.6} & \cellcolor{gray!20} \textbf{87.0} \\
Input Concat & 81.3 & 83.9 \\
Output Residual & 80.9 & 83.3 \\
\end{tabular}
\end{minipage}
} \\
\end{tabular}
\end{table*}

\begin{table*}[t]
\caption{Performance comparison with different modalities}
  \centering
  \renewcommand\arraystretch{1}
  \setlength\tabcolsep{17.2pt}

    \begin{tabular}{c|ccccccc}
      \noalign{\hrule height 1pt}
      \textbf{Modality} & \textbf{RGB} & \textbf{NIR} & \textbf{TIR} & \textbf{RGB+NIR} & \textbf{RGB+TIR} & \textbf{NIR+TIR} & \textbf{RGB+NIR+TIR} \\
      \hline
      mAP & 49.0 & 34.5 & 49.1 & 63.9 & 72.9 & 66.4 & 82.6 \\
      \noalign{\hrule height 1pt}
    \end{tabular}
  \label{tab:diffmodal}
\end{table*}

\begin{table}[t]
\caption{Ablation study on mask generation strategies
}
  \centering
  \setlength{\tabcolsep}{24.75pt}
  \begin{tabular}{l|cc}
    \noalign{\hrule height 1pt}
    \textbf{Mask Method} & \textbf{mAP} & \textbf{R-1}  \\ \hline
    RGB & 79.6	 & 82.5  \\
    Additive Fusion & 80.1 & 83.6  \\
    YCbCr Equal Weights & 82.0 & 84.9  \\
    \rowcolor[gray]{0.92}
    \textbf{YCbCr Fuse (Ours)} & \textbf{82.6} & \textbf{87.0}  \\
    \noalign{\hrule height 1pt}
  \end{tabular}
  \label{tab:mask}
\end{table}

\subsubsection{Multi-Modal Vehicle ReID}
Table~\ref{tab:multi-spectral vehicle ReID} reports the evaluation results on the RGBNT100 and MSVR310 datasets.
On the large-scale RGBNT100 benchmark, Ours$^\dagger$ achieves 89.4\% mAP, surpassing TOP-ReID$^*$ by 8.2\%. This significant gain demonstrates the effectiveness of our framework in leveraging multi-spectral complementarity for discriminative feature learning at scale.
On the more challenging MSVR310 dataset, which contains extreme lighting variations, heavy occlusions, and complex backgrounds, Ours$^\dagger$ obtains 64.6\% mAP and 76.0\% Rank-1 accuracy, yielding improvements of 17.6\% and 13.6\% over IDEA$^\dagger$, respectively.
{We note that this large gain is not an isolated outlier. Recent semantic-prior-based methods~\cite{li2025next} report similarly substantial improvements over IDEA on the same dataset (13.8\% mAP), suggesting that MSVR310 is intrinsically sensitive to semantic guidance.}
The substantial performance boost under such adverse conditions underscores the benefit of our dual semantic guidance. The structure-enhanced soft mask effectively suppresses background clutter, while the unified textual prior ensures consistent cross-modal grounding even when visual appearances are severely degraded.
These consistent improvements across datasets with distinct characteristics validate both the robustness and generalization capability of the proposed framework, positioning Ours$^\dagger$ as a strong contender among SOTA multi-modal vehicle ReID methods.

\subsection{Ablation Studies}
We evaluate the contribution of each component in our framework on the RGBNT201 dataset.

\subsubsection{Effect of Key Modules}
We conduct ablation studies to evaluate the contribution of the key modules: the Text-Semantic Injector (TSI), the Masked Global-Local Modulator (MGLM), and the Hierarchical MoE Fusion (HMF). 
Table~\ref{tab:main_ablation} presents the performance of different architectural variants.
Model A serves as the baseline, achieving an mAP of 70.3\% and a Rank-1 accuracy of 72.1\%. This variant uses only a shared CLIP encoder with naive feature concatenation, lacking explicit semantic guidance or structured modulation.
By incorporating TSI, Model B improves the mAP to 76.1\% and Rank-1 accuracy to 79.1\%, demonstrating the effectiveness of unified textual prompting in establishing consistent cross-modal semantic grounding. The gain confirms that aligning multi-spectral inputs through a shared language space mitigates modality-specific bias and validates the efficacy of high-dimensional hypergraph-based interaction between image and text features.
Model C further integrates MGLM, yielding an mAP of 78.8\% and Rank-1 accuracy of 84.0\%. This improvement stems from the structure-aware soft mask, which serves as a spatial semantic signal to enable foreground-focused local modeling. By dynamically modulating patch tokens based on spatial semantics, MGLM effectively suppresses background noise and aligns part-level correspondences across modalities, especially under conditions such as occlusion or low illumination.
Finally, Model D introduces the Hierarchical MoE Fusion (HMF), achieving the best performance with an mAP of 82.6\% and Rank-1 accuracy of 87.0\%. HMF replaces static fusion with adaptive, context-aware routing, allowing the model to selectively emphasize informative modalities at both global and local levels. This hierarchical design resolves the limitations of fixed-weight fusion and enables fine-grained cross-modal collaboration.
These results validate the incremental improvements contributed by each proposed module and collectively demonstrate that dual semantic guidance, global-local mutual modulation, and adaptive expert fusion are essential for robust multi-modal ReID.

\subsubsection{Computational Efficiency}
Computational efficiency is a critical metric for evaluating the practicality of ReID methods.
{Following the same reporting convention as existing MLLM-assisted ReID works~\cite{wang2025idea,li2025next,xu2026stmi}, FLOPs account for all forward-pass operations including the frozen CLIP text encoder (3.01\,G FLOPs), and training throughput is measured with textual descriptions and masks pre-generated and cached.}
Specifically, Table~\ref{tab:main_ablation} includes an ablation study on computational cost during inference.
The baseline model incurs 34.27\,G FLOPs, and each of our proposed modules adds only marginal overhead.
This is because the dominant computational burden comes from the image and text encoders, while TSI, MGLM, and HMF are designed to be lightweight.
In Table~\ref{tab:flops}, we further compare our method against state-of-the-art ViT- and CLIP-based approaches in terms of both FLOPs and training throughput.
Our full model achieves 37.91\,G FLOPs with a sampling rate of 169.40 samples per second.
The throughput is comparable to that of methods with similar FLOPs, as the computational bottleneck remains the shared encoders rather than our lightweight modules.
Moreover, as shown in Table~\ref{tab:params}, our model achieves competitive performance with a reasonable number of trainable parameters, indicating a trade-off between accuracy and model complexity.

The above FLOPs and throughput figures do not include the offline preprocessing stage.
For GPT-4o, its exact FLOPs cannot be measured as it is accessed through a closed-source API.
We instead quantify the practical offline latency and cache storage in Table~\ref{tab:offline_preprocess_overhead}. SAM2 mask generation requires 0.0115\,s per sample, and the total cache storage is modest (e.g., 1.16\,MB text + 3.45\,MB mask for RGBNT201).
These priors are loaded directly from disk during training and inference, incurring negligible runtime overhead.

\begin{table}[t]
    \centering
    \caption{Offline pre-processing overhead of the generated textual and mask priors}
    \label{tab:offline_preprocess_overhead}
    \renewcommand\arraystretch{1.18}
    \setlength{\tabcolsep}{7.2pt}
    \begin{tabular}{r|ccc}
        \noalign{\hrule height 1pt}
        \textbf{Metric} & \textbf{RGBNT201} & \textbf{RGBNT100} & \textbf{MSVR310} \\\hline
        SAM2 Time (s/sample) & 0.0115 & 0.0115 & 0.0115 \\
        Text Storage (MB)    & 1.16   & 4.68   & 0.55   \\
        Mask Storage (MB)    & 3.45   & 12.43  & 1.50   \\
        \noalign{\hrule height 1pt}
    \end{tabular}
\end{table}

\subsubsection{Component and Parameter Analysis of Core Modules}
We present an ablation study on the internal design choices of each core module.
In Table~\ref{tab:sub_ablation}, we analyze the internal designs of the Text-Semantic Injector (TSI), Masked Global-Local Modulator (MGLM), and Hierarchical MoE Fusion (HMF) modules, respectively.

The TSI module incorporates textual information and employs GeM pooling to extract a compact global text feature. 
Model A removes textual input entirely, while Model B retains the text encoder but removes GeM pooling. 
{Model C disables the padding mask $M_{\text{pad}}$, allowing padding (PAD) tokens to participate in self-attention. This variant achieves 82.1\% mAP and 85.0\% Rank-1 accuracy, yet still trails the full model, confirming that masking irrelevant padding tokens is necessary for clean textual control signals.}
Both variants underperform the full model, demonstrating that unified textual prompting provides essential cross-modal semantic grounding and that GeM pooling is crucial for distilling discriminative global descriptors from noisy language embeddings.

The MGLM module performs mask-guided trimodal local modeling and incorporates the fused global token $f^{\text{global}}$ to enable global-to-local modulation. 
We evaluate three variants:
{Model D} removes the structure-aware mask, resulting in a significant performance drop. This degradation occurs because the model can no longer leverage mask-based attention to focus on semantically critical regions.
{Model E} processes each modality independently without cross-modal interaction, yielding inferior results and confirming that explicit inter-modality collaboration is essential. This cross-modal design principle fundamentally distinguishes our approach from single-modality feature processing.
{Model F} excludes the global token $f^{\text{global}}$ during local modeling to verify the necessity of global awareness. The performance decline validates that the condensed global representation effectively guides cross-modal alignment at the patch level.
This effectiveness stems from our structure-preserving mask, which consistently highlights foreground parts across RGB, NIR, and TIR modalities, thereby reducing cross-modal misalignment in local features. 
Moreover, jointly modeling aligned trimodal patches while allowing local representations to access global context significantly improves performance, confirming that identity-discriminative cues emerge from the synergy between spatially coherent local details and holistic semantics.

Our HMF module adopts a two-level routing strategy: in the first stage, a Part-Aware MoE captures local context to modulate a global MoE in the second stage, enabling local-to-global feature refinement. 
Ablation studies confirm this design. Replacing the modulated MoE with a standard global MoE ({Model H}) degrades performance, as static expert weights fail to adapt to local structural variations. 
The part-only variant ({Model G}) achieves only 78.3\% mAP and performs worse than the full model due to insufficient global context.
These results highlight that mutual modulation between local details and global semantics, rather than isolated processing or naive fusion, is essential for discriminative multi-modal representation learning.

\subsubsection{Component and Hyperparameter Ablation}
We further analyze key hyperparameters and structural choices in Table~\ref{tab:ablations} to validate the robustness and optimality of our design.

To verify the impact of textual feature granularity, we test different numbers of GeM pooling branches for text feature extraction in Table~\ref{tab:ablations}(a). 
We observe that using two branches achieves the best trade-off between performance and computational complexity, yielding higher mAP and Rank-1 accuracy than both single-branch and three-branch variants. 
This indicates that a moderate level of feature diversification, which captures complementary semantic aspects without excessive redundancy, is optimal for cross-modal alignment. The performance drop with more branches suggests diminishing returns and potential overfitting due to increased parameters.

To assess the sensitivity of our hypergraph construction to sparsity, we vary the hypergraph sparsity threshold $\tau$ in Table~\ref{tab:ablations}(b). 
Setting $\tau=2$ consistently yields the best ReID performance across datasets. 
This value strikes an effective balance: it retains sufficient high-order node connections to model complex inter-part relationships while avoiding overly dense graphs that introduce noise and degrade generalization. The decline in performance when $\tau=1$ (too sparse) or $\tau \geq 3$ (too dense) confirms that our hypergraph-based interaction is highly sensitive to this threshold, and $\tau=2$ is indeed the optimal choice.

To justify our fusion architecture, we compare the Hypergraph-based Feature Refinement (HFR) module against standard Cross-Attention in Table~\ref{tab:ablations}(c). 
Our HFR significantly outperforms Cross-Attention in both mAP and Rank-1 metrics. 
This gain stems from HFR’s ability to model high-order, flexible interactions among multiple local regions simultaneously, whereas Cross-Attention is limited to pairwise token-level correlations. The result validates that moving beyond pairwise modeling is essential for capturing the holistic structure of multi-modal person representations.

To determine the most effective source of global guidance, we evaluate three strategies in Table~\ref{tab:ablations}(d): (i) RGB-only global token, (ii) weighted fusion of trimodal global tokens, and (iii) direct concatenation of all three. 
The RGB-only variant achieves the highest performance. 
We attribute this to the fact that TIR and NIR primarily convey structural cues at the patch level, while their global features, which are often contaminated by inaccurate color semantics or low-frequency artifacts, provide misleading guidance during local refinement. In contrast, the RGB global token offers rich, photorealistic semantic context that effectively steers cross-modal alignment. This finding aligns with our mask generation principle, where RGB serves as the chrominance anchor while TIR/NIR enhance structure, as illustrated in Fig.~\ref{fig:mask_y}.

To optimize the capacity of our Mixture-of-Experts routing, we sweep the number of experts $k$ in Table~\ref{tab:ablations}(e). 
Setting $k=4$ delivers significant performance gains with only a modest increase in trainable parameters, demonstrating high parameter efficiency. 
Further increasing $k$ leads to marginal improvements at the cost of higher memory and computation, indicating that four experts are sufficient to capture the necessary modality- and part-specific patterns without over-parameterization.

Finally, in Table~\ref{tab:ablations}(f), we compare different interaction strategies within HMF, including gate modulation, input concatenation, and output residual.
Our approach achieves the best results, confirming that adaptive, mask-aware modulation of local features is critical for foreground-focused representation learning. The performance gap highlights the contribution of our guidance mechanism in suppressing background interference and enhancing cross-modal consistency.

Moreover, across all ablation studies, the model performance varies smoothly and predictably with respect to hyperparameter changes, without abrupt fluctuations. This consistent trend demonstrates that our method is stable and not overly sensitive to hyperparameter tuning, which is crucial for practical deployment in real-world multi-modal ReID scenarios.

We evaluate various modality configurations, including single-modal (RGB, NIR, TIR) and multi-modal inputs, as summarized in Table~\ref{tab:diffmodal}.
On the multi-modal ReID benchmark, our method outperforms all single-modality baselines~\cite{sun2018beyond,zhou2019omni,rao2021counterfactual,he2021transreid}, confirming that fusing complementary spectral cues enhances robustness and discriminability.

\subsubsection{Analysis of Mask Generation Strategies}
Since the quality of the generated mask significantly affects segmentation-based ReID performance, we further evaluate different mask generation strategies. 
Specifically, we compare masks produced from the original RGB images, those from directly fused RGB-NIR-TIR via weighted summation, and those derived from our YCbCr-fused image. 
{As shown in Table~\ref{tab:mask}, YCbCr Equal Weights (uniform $1/3$) further improves to 82.0\% mAP and 84.9\% Rank-1, validating luminance-space fusion. 
Our asymmetric RGB-dominant fusion achieves the best results (82.6\% mAP / 87.0\% Rank-1). 
The marginal gap between YCbCr Equal Weights and our fixed weights (0.6\% mAP) indicates that the YCbCr fusion framework itself is the primary contributor, while the specific coefficient assignment serves as a stable physical prior rather than a sensitive hyperparameter.}

\subsubsection{Robustness under Degraded Conditions}
To comprehensively evaluate the robustness of our framework, we conduct dedicated experiments under two challenging scenarios on the RGBNT201 dataset.
For partial occlusion, we simulate realistic blocking by applying rectangular masks at random spatial positions across all three modalities with area ratios varying from 0 to 0.4.
To model modality misalignment, we designate RGB as the reference and apply random horizontal and vertical shifts to NIR and TIR with magnitudes ranging from 0 to 50 pixels, emulating registration errors in practical multi-spectral capture systems.

\begin{table}[t]
  \caption{Performance comparison (mAP) under different pixel-level modality misalignment. The column headers indicate the degree of random spatial shift (in pixels) applied to the auxiliary modality}
  \label{tab:pixel_offset}
  \centering
  \renewcommand\arraystretch{1.18}
  \setlength{\tabcolsep}{7.5pt}
  \begin{tabular}{r|cccccc}
      \noalign{\hrule height 1pt}
      \textbf{Methods} & 0 & 10 & 20 & 30 & 40 & 50 \\\hline
      PromptMA$^\dagger$~\cite{zhang2025prompt} & 78.4 & 78.1 & 77.2 & 74.2 & 70.8 & 66.1 \\
      DeMo$^\dagger$~\cite{wang2025decoupled} & 79.0 & 79.4 & 78.2 & 75.1 & 71.3 & 65.4 \\
      IDEA$^\dagger$~\cite{wang2025idea} & 80.2 & 81.3 & 79.5 & 76.7 & 71.4 & 66.1 \\
      Signal$^\dagger$~\cite{liu2026signal} & 80.3 & 80.0 & 79.5 & 78.5 & 76.6 & 72.9 \\
    \rowcolor[gray]{0.92} $\textbf{Ours}^\dagger$ & 82.6 & 82.5 & 81.5 & 80.7 & 77.5 & 73.2 \\
      \noalign{\hrule height 1pt}
  \end{tabular}
\end{table}

\begin{table}[t]
  \caption{Performance comparison (mAP) under different levels of image occlusion. The column headers indicate the ratio of the occluded area relative to the total image area}
  \label{tab:occluded_area}
  \centering
  \renewcommand\arraystretch{1.18}
  \setlength{\tabcolsep}{10.0pt}
  \begin{tabular}{r|ccccc}
      \noalign{\hrule height 1pt}
      \textbf{Methods} & 0 & 0.1 & 0.2 & 0.3 & 0.4 \\\hline
      PromptMA$^\dagger$~\cite{zhang2025prompt} & 78.4 & 74.3 & 64.8 & 58.4 & 45.5 \\
      DeMo$^\dagger$~\cite{wang2025decoupled} & 79.0 & 73.5 & 65.1 & 61.4 & 48.0 \\
      IDEA$^\dagger$~\cite{wang2025idea} & 80.2 & 76.1 & 68.2 & 63.9 & 49.8 \\
      Signal$^\dagger$~\cite{liu2026signal} & 80.3 & 75.4 & 67.1 & 59.8 & 49.0 \\
    \rowcolor[gray]{0.92} $\textbf{Ours}^\dagger$ & 82.6 & 78.4 & 71.8 & 66.6 & 55.2 \\
      \noalign{\hrule height 1pt}
  \end{tabular}
\end{table}

\begin{table}[t]
  \caption{Performance comparison (R-1) under different pixel-level modality misalignment. The column headers indicate the degree of random spatial shift (in pixels) applied to the auxiliary modality}
  \label{tab:pixel_offset_r1}
  \centering
  \renewcommand\arraystretch{1.18}
  \setlength{\tabcolsep}{7.5pt}
  \begin{tabular}{r|cccccc}
      \noalign{\hrule height 1pt}
      \textbf{Methods} & 0 & 10 & 20 & 30 & 40 & 50 \\\hline
      PromptMA$^\dagger$~\cite{zhang2025prompt} & 80.9 & 81.0 & 80.6 & 80.4 & 78.6 & 78.0 \\
      DeMo$^\dagger$~\cite{wang2025decoupled} & 82.3 & 81.8 & 81.0 & 80.3 & 78.8 & 77.3 \\
      IDEA$^\dagger$~\cite{wang2025idea} & 82.1 & 84.8 & 84.6 & 81.7 & 79.2 & 78.1 \\
      Signal$^\dagger$~\cite{liu2026signal} & 85.2 & 85.9 & 83.9 & 83.5 & 80.6 & 79.1 \\
    \rowcolor[gray]{0.92} $\textbf{Ours}^\dagger$ & 87.0 & 87.1 & 85.8 & 84.9 & 81.3 & 79.3 \\
      \noalign{\hrule height 1pt}
  \end{tabular}
\end{table}

\begin{table}[t]
  \caption{Performance comparison (R-1) under different levels of image occlusion. The column headers indicate the ratio of the occluded area relative to the total image area}
  \label{tab:occluded_area_r1}
  \centering
  \renewcommand\arraystretch{1.18}
  \setlength{\tabcolsep}{10.0pt}
  \begin{tabular}{r|ccccc}
      \noalign{\hrule height 1pt}
      \textbf{Methods} & 0 & 0.1 & 0.2 & 0.3 & 0.4 \\\hline
      PromptMA$^\dagger$~\cite{zhang2025prompt} & 80.9 & 79.1 & 70.7 & 67.7 & 61.2 \\
      DeMo$^\dagger$~\cite{wang2025decoupled} & 82.3 & 77.6 & 70.3 & 68.1 & 65.0 \\
      IDEA$^\dagger$~\cite{wang2025idea} & 82.1 & 80.9 & 74.4 & 72.8 & 65.7 \\
      Signal$^\dagger$~\cite{liu2026signal} & 85.2 & 83.0 & 76.7 & 72.4 & 65.0 \\
    \rowcolor[gray]{0.92} $\textbf{Ours}^\dagger$ & 87.0 & 83.7 & 78.3 & 73.4 & 67.8 \\
      \noalign{\hrule height 1pt}
  \end{tabular}
\end{table}

\begin{figure}[t]
  \centering
    \includegraphics[width=\linewidth]{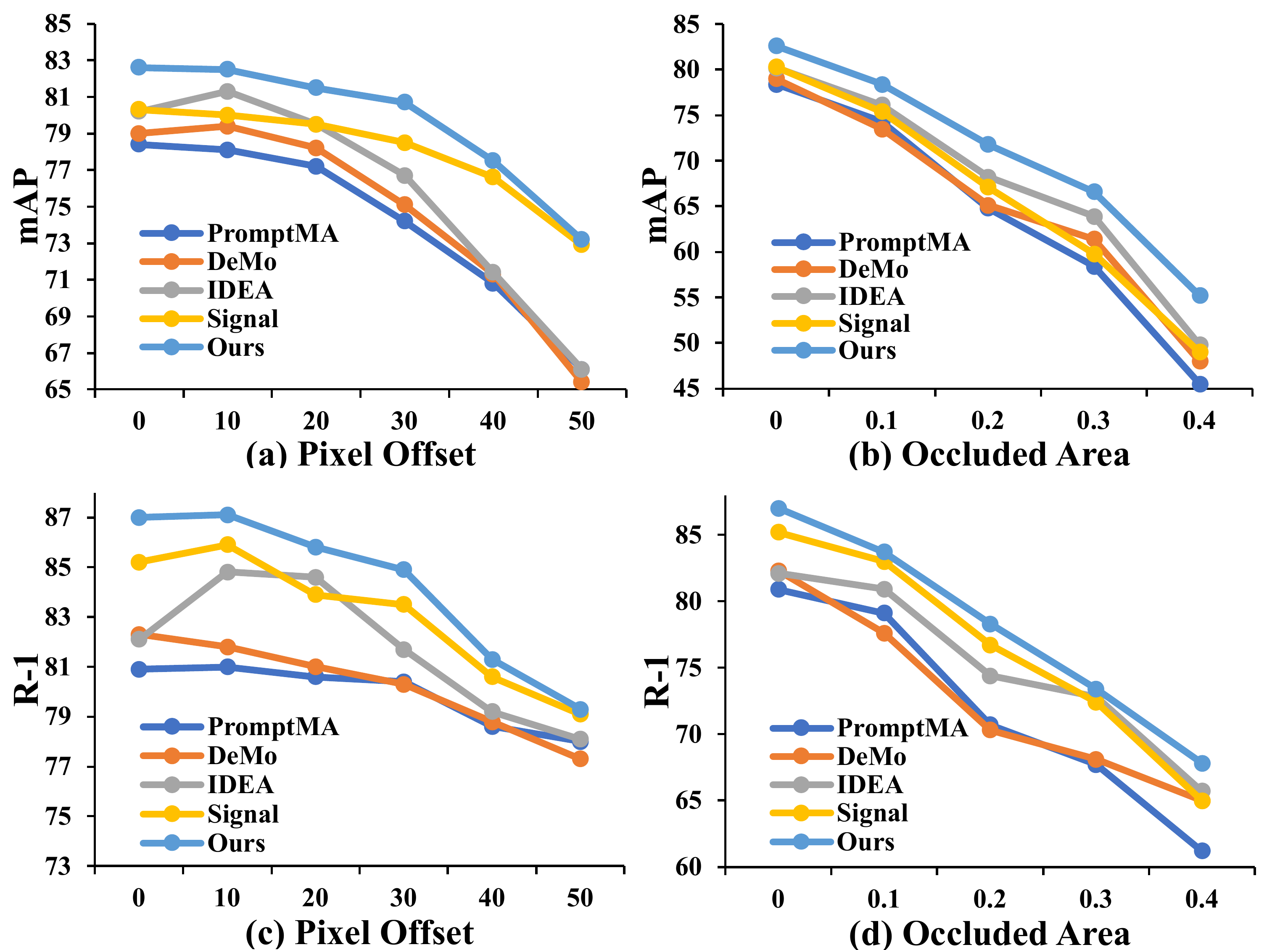}%
  \caption{Robustness evaluation on the RGBNT201 dataset under challenging conditions. Performance degradation with respect to (a), (c) pixel-level modality misalignment (spatial shift) and (b), (d) image occlusion (occluded area ratio), measured by mAP in (a), (b) and Rank-1 accuracy in (c), (d). The proposed method consistently outperforms across all evaluated scenarios.}
  \label{fig:robustness}
\end{figure}

\begin{figure*}
  \centering
  \includegraphics[width=\linewidth]{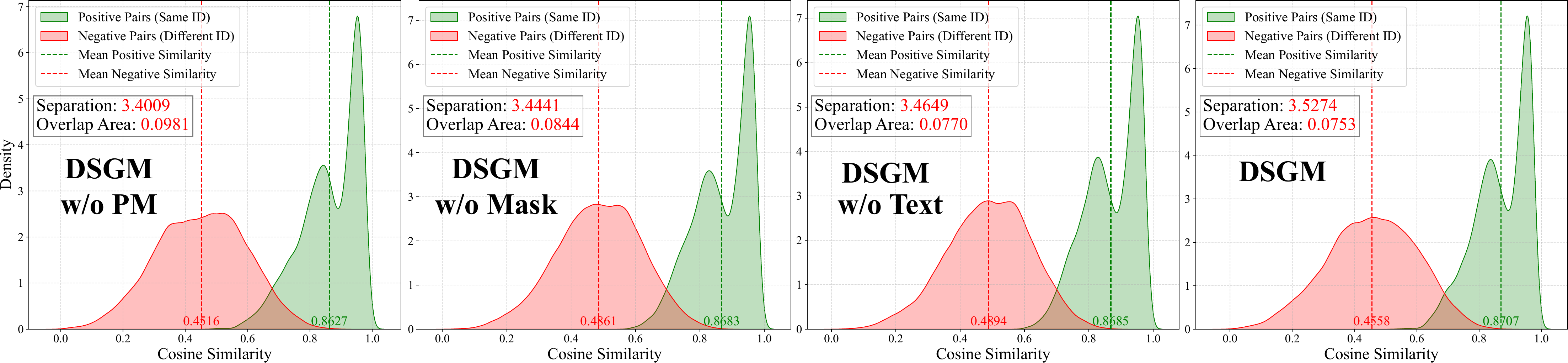}
  \vspace{-7mm}
   \caption{
    Cosine similarity distribution across identity classes. 
    The proposed method with Part-Aware MoE, mask, and text guidance achieves better class separation and reduced intra-class overlap, indicating enhanced discriminative feature learning through progressive interaction. 
    This provides evidence for the effectiveness of our two-stage routing mechanism and dual semantic modulation design.
    }
  \label{fig:cosine}
\end{figure*}

\begin{figure*}[t]
  \centering
  \includegraphics[width=0.97\linewidth]
  {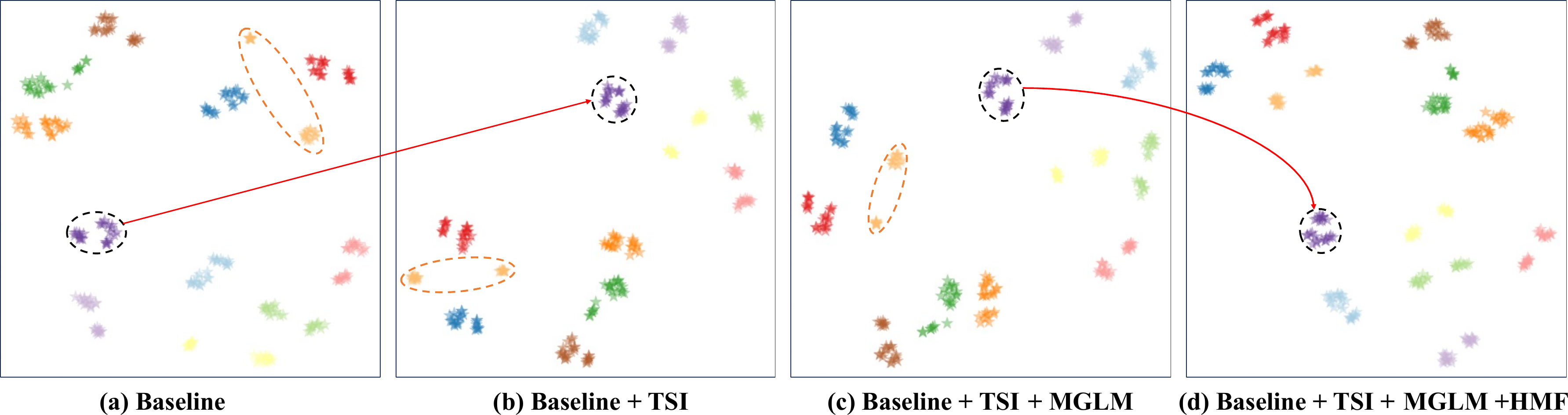}
  \vspace{-3mm}
   \caption{
    t-SNE visualization~\cite{van2008visualizing} of feature distributions. 
    Different colors represent different identities, showing the clustering and separability achieved by our method. 
    As key components are progressively added, features of the same identity become more compact while different identities are better separated, 
    demonstrating the incremental benefit of the proposed modules.
    }
  \label{fig:tsne}
\end{figure*}

As summarized in Tables~\ref{tab:pixel_offset}, \ref{tab:occluded_area}, \ref{tab:pixel_offset_r1}, and~\ref{tab:occluded_area_r1}, our method demonstrates consistently stable performance under both degradation types.
Notably, while Signal~\cite{liu2026signal} achieves competitive results under clean conditions (80.3\% mAP), our method exhibits superior robustness when data quality is compromised: under 50-pixel displacement, we maintain 73.2\% mAP versus 72.9\% for Signal, and under 40\% occlusion, we achieve 55.2\% mAP versus 49.0\%.
An interesting observation from the misalignment study is that a small offset (e.g., 10 pixels) leads to minor mAP fluctuations across methods: some exhibit slight improvement (e.g., DeMo from 79.0\% to 79.4\%), while others show negligible variation (e.g., ours from 82.6\% to 82.5\%).
We attribute this to the centered composition of targets in RGBNT201, where minor shifts effectively reduce background clutter and act as implicit data augmentation.
The superior occlusion tolerance of our method originates from the explicit spatial priors embedded in our framework: the structure-aware soft masks guide MGLM to concentrate on visible foreground regions while attenuating occluded areas, and the hierarchical MoE fusion in HMF adaptively redistributes modality weights when certain spectral channels are compromised.
Fig.~\ref{fig:robustness} visualizes the complete degradation curves for both conditions.

\subsubsection{Cross-Dataset Generalization}
To assess the transferability of our learned representations against domain shift, we conduct cross-dataset evaluation by training on RGBNT100 and directly evaluating on MSVR310 via feature-based retrieval.
As shown in Table~\ref{tab:cross_dataset}, our method achieves 29.6\% mAP and 41.8\% Rank-1, substantially outperforming Signal (10.5\%/26.7\%) under the identical protocol.
This demonstrates that our multi-modal fusion strategy learns transferable feature representations when facing unseen complex scenarios characterized by heavy occlusion, low illumination, and diverse weather conditions.
Notably, this cross-dataset advantage is consistent with our superior same-dataset performance on both benchmarks (Table~\ref{tab:multi-spectral vehicle ReID}), confirming that our framework does not overfit to a specific dataset distribution.

\begin{table}[t]
    \centering
    \setlength{\tabcolsep}{5.3pt}
    \caption{Cross-dataset evaluation from RGBNT100 to MSVR310. The model is trained on the source dataset and directly evaluated on the target dataset by feature-based retrieval}
    \label{tab:cross_dataset}
    \begin{tabular}{l|cc|cc|cc}
        \noalign{\hrule height 1pt}
        \multirow{2}{*}{Method} & \multicolumn{2}{c|}{RGBNT100} & \multicolumn{2}{c|}{MSVR310} & \multicolumn{2}{c}{RGBNT100$\rightarrow$MSVR310} \\
        \cline{2-7}
        & mAP & R-1 & mAP & R-1 & mAP & R-1 \\
        \hline
        Signal$^\dagger$~\cite{liu2026signal} & 86.3 & 97.6 & 53.6 & 71.9 & 10.5 & 26.7 \\
        \textbf{Ours$^\dagger$} & \textbf{89.4} & \textbf{98.2} & \textbf{64.6} & \textbf{76.0} & \textbf{29.6} & \textbf{41.8} \\
        \noalign{\hrule height 1pt}
    \end{tabular}%
\end{table}

\subsubsection{Sensitivity to Text Generator Quality}
To verify that our framework does not critically depend on the GPT-4o API, we evaluate performance using Qwen-VL~\cite{bai2023qwen}, an open-source MLLM, to generate textual descriptions.
Compared with GPT-4o, Qwen-VL produces noisier text with occasional attribute recognition failures (e.g., incorrect clothing color or vehicle type).
As shown in Table~\ref{tab:weak_mllm}, when training and testing with these weaker textual descriptions, our method achieves 81.8\% mAP and 84.4\% Rank-1 on RGBNT201.
This represents only a modest drop of 0.8\% mAP and 2.6\% Rank-1 relative to using GPT-4o-generated text (82.6\% / 87.0\%), indicating that our framework is relatively robust to text quality degradation.
The modest degradation can be attributed to the design of TSI, where the padding mask suppresses noisy token interference and the hypergraph-based refinement filters out inconsistent semantic signals during vision-language interaction.

\begin{table}[t]
\centering
\caption{Performance of our method on RGBNT201 using different text generation models. Qwen-VL produces weaker text with occasional attribute recognition failures compared with GPT-4o}
\label{tab:weak_mllm}
\renewcommand\arraystretch{1.18}
\setlength{\tabcolsep}{16.5pt}
\begin{tabular}{l|cc|cc}
\noalign{\hrule height 1pt}
\multirow{2}{*}{Method} & \multicolumn{2}{c|}{w/ Qwen-VL Text} & \multicolumn{2}{c}{w/ GPT-4o Text} \\
\cline{2-5}
& mAP & R-1 & mAP & R-1 \\
\hline
$\textbf{Ours}^\dagger$ & 81.8 & 84.4 & 82.6 & 87.0 \\
\noalign{\hrule height 1pt}
\end{tabular}
\end{table}

\begin{figure}
  \centering
  \includegraphics[width=\linewidth]{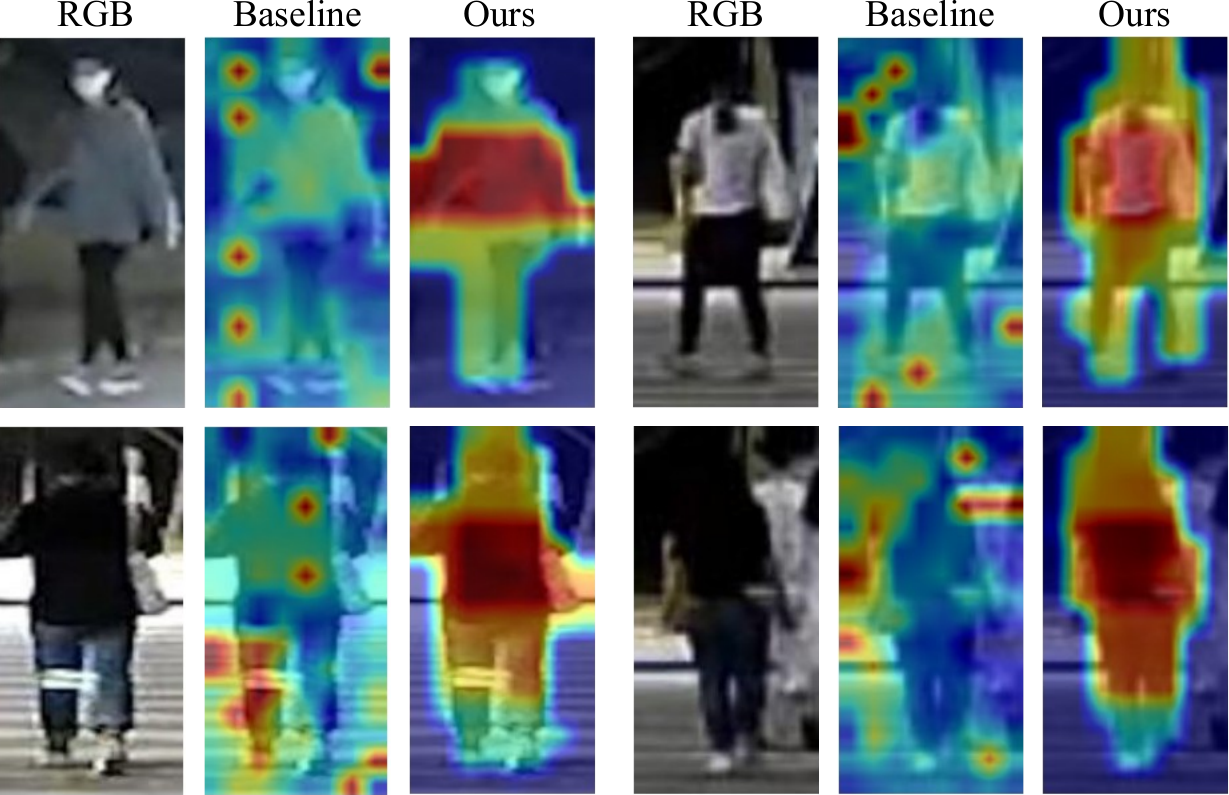}
   \caption{
    Visualization of channel activation maps for person ReID. 
    Each group compares the baseline (middle) and the proposed method (right), given the input image (left). 
    }
    \label{fig:map}
\end{figure}

\begin{figure}
  \centering
  \includegraphics[width=\linewidth]{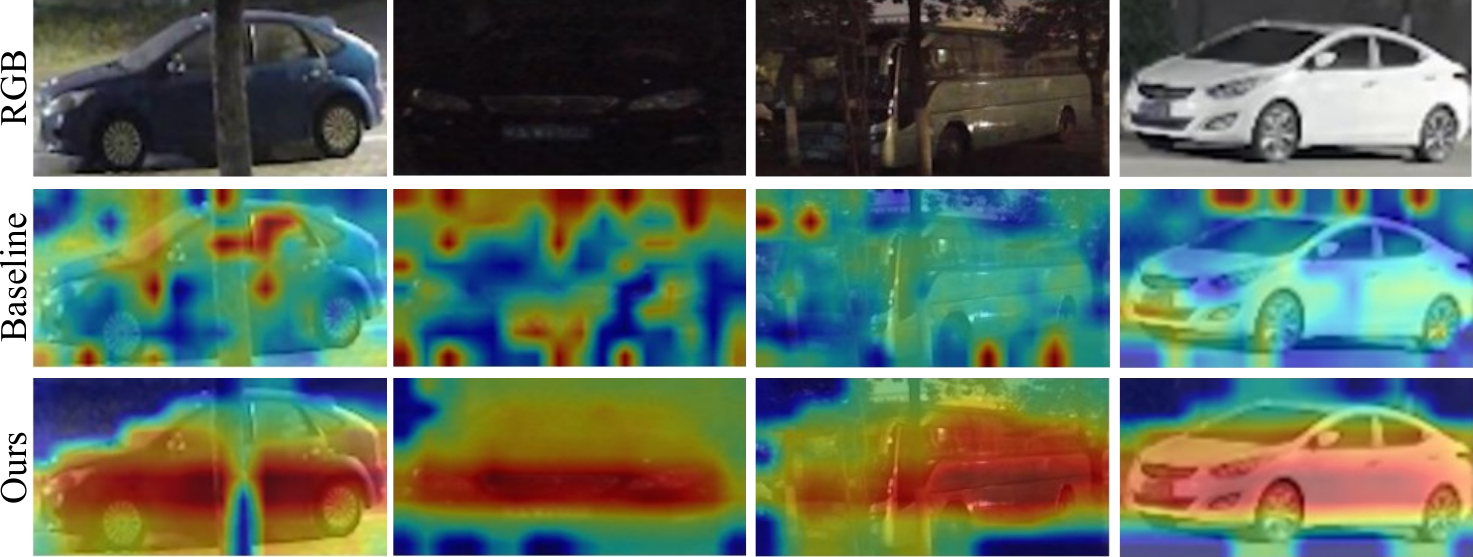}
   \caption{
    Visualization of channel activation maps for vehicle ReID. 
    Each group compares the baseline and the proposed method, given the input image. 
    }
    \label{fig:map_car}
\end{figure}

\begin{figure*}
  \centering
  \includegraphics[width=\linewidth]{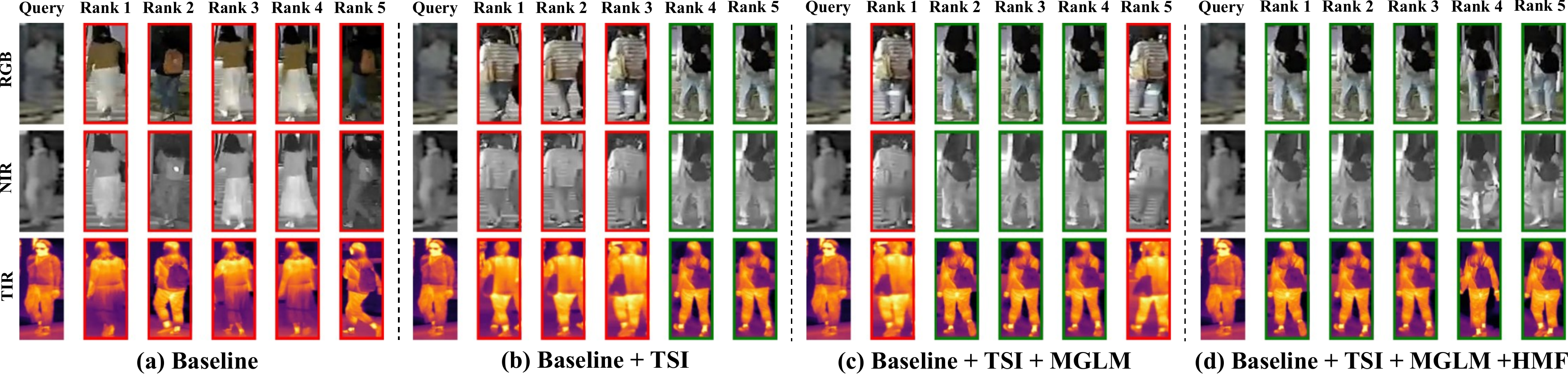}
   \caption{
    Top-5 retrieval results on a person ReID benchmark under progressive module integration. 
    Each sub-figure illustrates the impact of incrementally adding a core component to the baseline, 
    showing consistent gains in identity discrimination and robustness to pose, illumination, and occlusion variations.
}
  \label{fig:rank_list_p}
\end{figure*}

\begin{figure*}
  \centering
  \includegraphics[width=\linewidth]{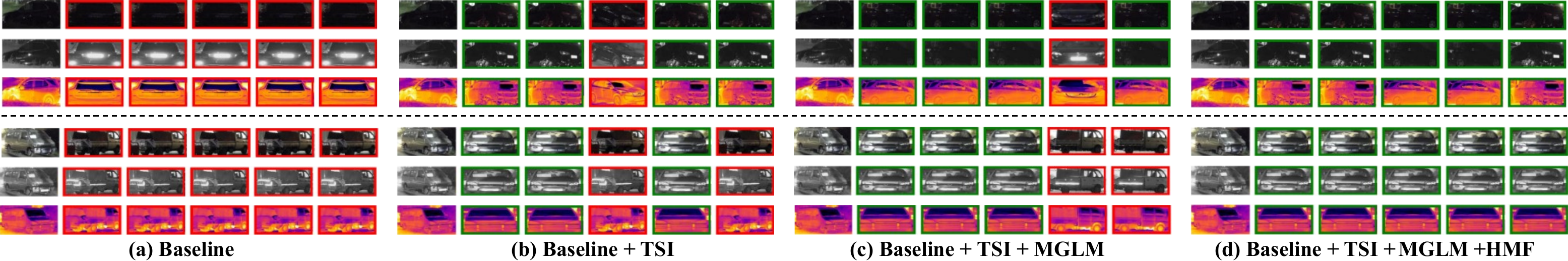}
   \caption{
    Top-5 retrieval results on a vehicle ReID benchmark under progressive module integration. 
    Each sub-figure illustrates the impact of incrementally adding a core component to the baseline, 
    demonstrating improved robustness to viewpoint changes, lighting conditions, and fine-grained appearance variations such as car model and color.
}
  \label{fig:rank_list_c}
\end{figure*}

\subsection{Visualization Analysis}

\subsubsection{Cosine Similarity Distributions}
To evaluate the impact of dual semantic priors and local modulation mechanisms on feature discriminability, we visualize the cosine similarity distributions between positive and negative test sample pairs, comparing the full model against ablated variants: w/o Part-Aware MoE (w/o PM), w/o mask guidance (w/o Mask), and w/o textual guidance (w/o Text), as shown in Fig.~\ref{fig:cosine}.

To quantitatively assess inter-class separability and intra-class compactness, we adopt two complementary metrics. First, we compute the standardized mean difference as a separation metric:
\begin{equation}
    \text{Separation Metric} = (\bar{X}_1 - \bar{X}_2) / \sqrt{(s_1^2 + s_2^2) / 2},
\end{equation}
where $\bar{X}_1$ and $\bar{X}_2$ denote the means of the positive and negative similarity distributions, and $s_1^2$, $s_2^2$ their variances. 
Second, to estimate distribution overlap, we apply Kernel Density Estimation (KDE) and compute the intersection area:
\begin{equation}
    \text{Overlap Area} \approx \int \min(f_1(x), f_2(x))  dx,
\end{equation}
with $f_1(x)$ and $f_2(x)$ being the KDE-estimated densities for positive and negative samples, respectively.
These two metrics jointly offer a comprehensive evaluation of feature discriminability. Higher separation and lower overlap indicate better clustering and generalization.
The results show that the full model achieves the highest Separation Metric (3.5274) and the lowest Overlap Area (0.0753), indicating superior clustering structure. In contrast, ablating any component, particularly the mask or textual guidance, leads to increased distribution overlap and reduced separation, demonstrating their critical role in enhancing feature discriminability.

\subsubsection{Multi-Modal Feature Distributions}
To analyze how individual modules shape the global feature geometry across modalities, we employ t-SNE~\cite{van2008visualizing} to visualize embeddings from different model variants, as shown in Fig.~\ref{fig:tsne}.
Compared to the baseline in Fig.~\ref{fig:tsne}(a), the integration of TSI and MGLM (Fig.~\ref{fig:tsne}(b)–(c)) progressively refines ambiguous regions, as highlighted in the orange circle, by reducing inter-class overlap and sharpening cluster boundaries. 
Furthermore, adding the HMF module (Fig.~\ref{fig:tsne}(d)) yields even tighter intra-class clustering, suggesting its effectiveness in harmonizing multi-granularity features. 
These observations confirm that each component contributes incrementally to both inter-class separability and intra-class consistency in the embedding space.

\subsubsection{Visualization of Channel Activation Maps}
To investigate how our design influences spatial attention and semantic focus, we visualize channel activation maps after the MGLM module for both person and vehicle queries, as shown in Fig.~\ref{fig:map} and Fig.~\ref{fig:map_car}.
Our method, equipped with dual semantic guidance and global-to-local modulation, consistently activates coherent, object-centric regions, even under low illumination or occlusion. 
In contrast, the baseline exhibits diffuse or background-dominated responses, indicating weaker semantic grounding. 
This comparison demonstrates that explicit local modeling guided by masks and text steers the network toward more interpretable and robust feature representations.

\subsubsection{Rank List Comparison}
To qualitatively assess retrieval performance under real-world challenges, we compare the top-ranked results from the baseline and progressively enhanced variants on cross-camera person and vehicle queries, as shown in Fig.~\ref{fig:rank_list_p} and Fig.~\ref{fig:rank_list_c}.
The full model consistently retrieves true matches at higher ranks, even for low-quality or pose-varied queries. 
By contrast, the baseline and ablated models suffer from false positives and ranking instability, particularly when key components such as MGLM or HMF are removed. 
This progression validates that each module contributes complementary improvements in retrieval precision and robustness, collectively enabling reliable cross-modal matching.

\section{Conclusion}
\label{sec:conclusion}
In this work, we present a novel dual semantic-guided framework for multi-modal object ReID. 
To enrich semantic supervision and extend existing datasets, we leverage a pre-trained mask generator and MLLMs to produce instance-aligned masks and unified textual descriptions across modalities. 
Based on this, we design the Text-Semantic Injector (TSI), which enables interference-free text feature extraction and establishes high-order interactions between visual and textual modalities through hypergraph-based refinement. 
The Masked Global-Local Modulator (MGLM) introduces a mask-guided architecture for cross-modal local modeling, allowing local tokens to access global context while focusing on semantically relevant regions. 
Furthermore, the Hierarchical MoE Fusion (HMF) module implements a local-to-global, mixture-of-experts-based fusion strategy, dynamically modulating trimodal features in a context-aware manner. 
Extensive experiments on three public multi-modal ReID benchmarks demonstrate the superiority of our approach over state-of-the-art methods, validating the effectiveness of our dual semantic guidance and hierarchical modulation design.

\bibliographystyle{IEEEtran}
\bibliography{reference}

\vfill

\end{document}